\documentclass{article} 
\usepackage{Style/iclr2027_conference,times}

\usepackage{Style/iclr2027_conference,times}

\iclrfinalcopy

\usepackage{microtype}
\usepackage{inconsolata}
\usepackage{latexsym}

\usepackage{amsmath}
\usepackage{amssymb}
\usepackage{bm}

\usepackage{graphicx}
\usepackage{subcaption}
\usepackage{rotating}

\usepackage[table]{xcolor}
\usepackage{booktabs}
\usepackage{multirow}
\usepackage{makecell}
\usepackage{diagbox}
\usepackage{array}
\usepackage{tabularx}
\usepackage{longtable}
\usepackage{adjustbox}
\usepackage{ragged2e}

\usepackage{enumitem}
\usepackage{soul}
\usepackage{multicol}
\setlist{nosep}

\usepackage{algorithm}
\usepackage{algpseudocode}

\usepackage{placeins}
\usepackage{fontawesome}
\usepackage[most]{tcolorbox}
\usepackage{pgf}

\usepackage{url}
\usepackage{hyperref}

\newcommand{\imp}[1]{%
  \textcolor{blue!90!black}{\scalebox{0.72}{\,(+#1)}}%
}

\newcommand{\dec}[1]{%
  \textcolor{red!90!black}{\scalebox{0.72}{\,(-#1)}}%
}

\newcommand{\globalbest}[1]{\cellcolor{blue!15}\textbf{#1}}
\newcommand{\bestone}[1]{\cellcolor{blue!25}#1}
\newcommand{\besttwo}[1]{\cellcolor{blue!16}#1}
\newcommand{\bestthree}[1]{\cellcolor{blue!8}#1}

\newcommand{\worstone}[1]{\cellcolor{orange!30}#1}
\newcommand{\worsttwo}[1]{\cellcolor{orange!15}#1}

\newcommand{\impcell}[2]{%
    \pgfmathtruncatemacro{\shadevalue}{min(45,max(6,6 + 1.5*(#2)))}%
    \edef\tempcolor{\noexpand\cellcolor{blue!\shadevalue!white}}%
    \tempcolor #1%
}

\newcommand{\deccell}[2]{%
    \pgfmathtruncatemacro{\shadevalue}{min(55,max(6,6 + 1.1*(#2)))}%
    \edef\tempcolor{\noexpand\cellcolor{orange!\shadevalue!white}}%
    \tempcolor #1%
}

\newcommand{\scorecell}[1]{%
    \pgfmathtruncatemacro{\shadevalue}{max(4,min(55,58 - 0.58*(#1)))}%
    \edef\tempcolor{\noexpand\cellcolor{orange!\shadevalue!white}}%
    \tempcolor #1%
}

\newcommand{\gaincell}[1]{%
    \cellcolor{blue!15}#1%
}

\newcommand{\drop}[1]{\textcolor{red!80!black}{\tiny (#1)}}
\newcommand{\bigdrop}[1]{\textcolor{red!90!black}{\tiny\textbf{(#1)}}}

\usepackage[most]{tcolorbox}
\usepackage{enumitem}
\usepackage{xcolor}

\definecolor{promptgray}{RGB}{248,248,248}
\definecolor{promptborder}{RGB}{190,190,190}

\newtcolorbox{promptbox}[1]{
  breakable,
  colback=promptgray,
  colframe=promptborder,
  boxrule=0.5pt,
  arc=1mm,
  left=2mm,
  right=2mm,
  top=1.5mm,
  bottom=1.5mm,
  title=\textbf{#1},
  fonttitle=\small,
  fontupper=\small,
  before skip=5pt,
  after skip=5pt
}

\usepackage{xcolor}

\definecolor{overlapcolor}{RGB}{0,100,70}
\definecolor{conflictcolor}{RGB}{180,40,40}
\definecolor{unique1color}{RGB}{35,75,160}
\definecolor{unique2color}{RGB}{105,35,140}

\definecolor{overlapbg}{RGB}{220,240,232}
\definecolor{conflictbg}{RGB}{248,225,225}
\definecolor{unique1bg}{RGB}{225,232,248}
\definecolor{unique2bg}{RGB}{238,225,245}

\newcommand{\annbox}[2]{%
  \begingroup
  \setlength{\fboxsep}{0.25em}%
  \colorbox{#1}{#2}%
  \endgroup
}

\newcommand{\ov}[1]{%
  \annbox{overlapbg}{\textcolor{overlapcolor}{\textbf{O#1}}}%
}

\newcommand{\cf}[1]{%
  \annbox{conflictbg}{\textcolor{conflictcolor}{\textbf{C#1}}}%
}

\newcommand{\ua}[1]{%
  \annbox{unique1bg}{\textcolor{unique1color}{\textbf{U$_A$-#1}}}%
}

\newcommand{\ub}[1]{%
  \annbox{unique2bg}{\textcolor{unique2color}{\textbf{U$_B$-#1}}}%
}

\definecolor{domainbg}{RGB}{235,235,235}

\title{Overlap, Unique and Conflict: Can LLMs Extract What They Can Recognize?}

\author{
Eftekhar Hossain \qquad Santu Karmaker \\
Bridge-AI Lab@UCF, Department of Computer Science \\
University of Central Florida, USA \\
\texttt{\{eftekhar,santu\}@ucf.edu}
}

\begin{document}

\maketitle

\begin{abstract}
Understanding multi-perspective alternative narratives requires identifying how their information agrees, conflicts, or differs across sources. Existing work on cross-text relations largely focuses on categorizing relations between predefined text pairs, such as entailment or contradiction, rather than directly extracting such information from full narratives. To address this gap, we introduce \textbf{\textit{Overlap--Unique--Conflict (OUC)}} extraction, a cross-narrative task that extracts all overlapping, conflicting, and unique clauses from two narratives. To support this study, we construct a benchmark of approximately 22K narrative pairs and 140K OUC instances spanning factual, argumentative, and political discourse. Evaluating 14 open-source LLMs (0.6B--35B), we find that unique information is far easier to extract than overlap and conflict: the strongest model, Gemma-4-31B, reaches only 61.13\% F1-score on overlap and 48.58\% on conflict, against more than 75\% on unique. Further diagnostic analysis reveals that this difficulty does not stem from relation recognition alone, but rather from a failure to pair and extract the corresponding clauses from full narratives, especially in smaller models. Nevertheless, learning these extractions with task-specific supervision narrows the gap considerably: a fine-tuned Qwen-3-8B gains $\sim\textbf {15--28\%} $ absolute over its baseline and surpasses models roughly four times its size (e.g., Qwen-3.6-35B) on several tasks. Even so, overlap and conflict remain well below satisfactory, leaving cross-narrative clause extraction an open challenge. The benchmark is available at \url{https://huggingface.co/datasets/BridgeAI-Lab/OUC-Benchmark}.
\end{abstract}

\section{Introduction}
Information about the same event or topic is often distributed across multiple narratives rather than contained in a single source. While these narratives may agree on many details, they can also emphasize different aspects, contradict one another, or contain information that appears nowhere else. This makes it important to understand how the information in one narrative relates to that in another, especially in applications such as news analysis, peer-review synthesis, and incident reporting. While existing NLP tasks capture parts of this problem, they usually assume that the information to be compared is already known. Natural language inference (NLI), for example, determines the relation between a given premise--hypothesis pair \citep{bowman-etal-2015-large,williams-etal-2018-broad}, while fact verification begins with a predefined claim and retrieves evidence to support or refute it \citep{thorne-etal-2018-fever}. In both cases, the information to be examined is specified in advance. A similar assumption also appears in recent long-context and multi-document evaluations of LLMs \citep{zhu-etal-2024-fanoutqa,wang-etal-2024-leave,bai-etal-2025-longbench}, where models are still guided by a question or other explicit information need. Therefore, it remains unclear whether LLMs can identify and extract all shared, conflicting, and unique information directly from complete narratives without such guidance.
\begin{figure}
\vspace{-4mm}
    \centering
    \includegraphics[width=\textwidth]{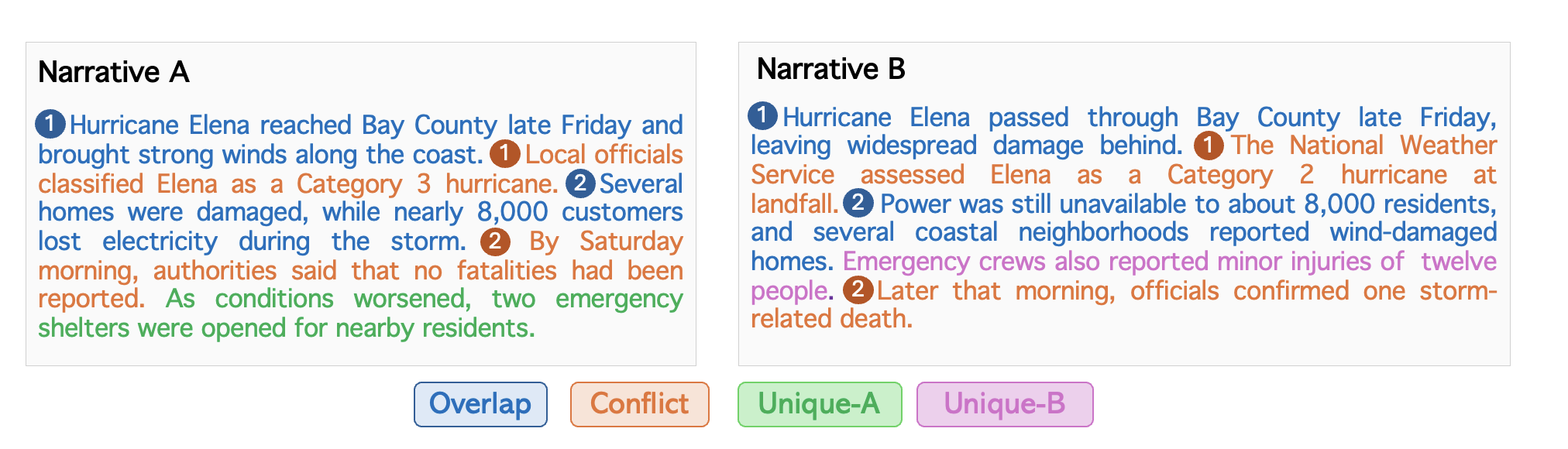}
    \vspace{-6mm}
    \caption{Illustrative example of Overlap, Conflict, and Unique information across two narratives. Highlighted spans denote the clauses to be extracted, while numbered links indicate the corresponding Overlapping and Conflicting pairs.}
    \label{fig:illustration}
    \vspace{-5mm}
\end{figure}

To address this gap, we introduce \textbf{\textit{Overlap--Unique--Conflict (OUC)}} clause extraction, a new cross-narrative task in which the model receives two alternative narratives and extracts all overlapping, conflicting, and narrative-specific unique information. Unlike pairwise inference, where the text units to be compared are provided in advance, OUC extraction starts from complete narratives and requires the relevant cross-narrative pairs and narrative-specific information to be identified during extraction.  To operationalize this task, we develop a benchmark comprising approximately 22K narrative pairs and about 140K OUC instances across three discourse types: factual, argumentative, and political. Using this benchmark, we evaluate 14 open-source LLMs ranging from 0.6B to 35B parameters under zero-shot, few-shot, and chain-of-thought instructions. Evaluations reveal that while current models can often recover information unique to a single narrative, they are less effective at extracting shared and conflicting information across narratives, particularly at smaller model scales (0.6B-8B). Surprisingly, this weakness does not reflect an equivalent inability to understand the relations themselves. Through a bottleneck analysis, we find that when the relevant sentence pairs are provided, even smaller models are much more successful at determining whether the pairs overlap or conflict. Their performance drops sharply when those pairs must be extracted from alternative narratives, revealing a substantial gap in current LLMs' cross-narrative extraction capabilities.

This motivates us to examine whether the gap observed in smaller models can be reduced by explicitly learning OUC extraction through task-specific supervision. Experiments with several forms of supervision (e.g., direct, preference-based, and reward-based) show that direct supervision yields the greatest gains. A learned 8B model, Qwen-3-8B, improves by roughly 19--28\% on the harder Overlap and Conflict tasks, and more than 15\% on both Unique tasks. More strikingly, the learned smaller model not only closes much of the gap with best-performing LLMs (e.g., Gemma-4-31B) but also surpasses models roughly four times its size (e.g., Qwen-3.6-35B) on conflict (+4.9\%) and unique tasks (about 7\% improvement). This suggests that OUC extraction performance is not determined by model scale alone, but can be learned effectively through task-specific supervision. We further find that these gains are not simply a consequence of using large-scale supervision, as much of the improvement can be achieved with only a fraction of the supervision size. Nevertheless, the gains do not fully resolve the problem, leaving accurate extraction of shared and conflicting information across narratives as an important open challenge. In summary, our main contributions are as follows:

\begin{itemize}[leftmargin=*,itemsep=0.4ex,partopsep=0.2ex,parsep=0ex]
    \item We introduce \textbf{Overlap--Unique--Conflict (OUC)} extraction, a new cross-narrative task to extract all overlapping, conflicting, and unique information between two alternative narratives. To support the study of this task, we develop a benchmark comprising approximately 22K narrative pairs and 140K OUC instances across factual, argumentative, and political discourse.
    \item We conduct an extensive evaluation across 14 open-source LLMs (0.6B--35B) to characterize how well current models perform OUC extraction and identify their main limitations. Our controlled analyses reveal a substantial gap between recognizing Overlap and Conflict from given pairs and extracting the same relations from complete narratives.
    \item We further demonstrate that task-specific supervision substantially improves OUC extraction in smaller LLMs, enabling an 8B model to match or outperform models roughly four times larger with only a fraction of the full supervision.
    
\end{itemize}

\section{Related Work}
\vspace{-1mm}
Cross-narrative understanding has long studied how information relates across texts describing the same event or topic. Early work on Cross-document Structure Theory modeled relations such as equivalence, elaboration, and contradiction between sentences from related documents~\citep{radev-2000-common,zhang2003learning,radev-etal-2004-cst,aleixo2008finding}. Other work has focused on aligning sentences or propositions that convey the same information despite differences in wording~\citep{nelken-shieber-2006-towards,grover-mitra-2017-sentence,weiss2021qa,molfese-etal-2024-neuralign}. More recent work has moved beyond sentence-level alignment toward structured cross-narrative extraction, including linking event mentions that refer to the same underlying event \citep{min-etal-2024-synergetic}, integrating event arguments distributed across multiple sources~\citep{gao-etal-2024-harvesting}, and extracting relations between entities whose supporting evidence is spread across documents~\citep{jain-etal-2024-knowledge,yue-etal-2024-towards}. Overall, these works provide different ways to connect information across documents, but they typically do so over predefined units such as events, arguments, or entity pairs.

A separate line of work has also examined semantic relations between such units through distinct NLP tasks. For example, paraphrase identification and semantic textual similarity have been used to capture shared or semantically equivalent information across texts~\citep{dolan-brockett-2005-automatically,agirre-etal-2012-semeval,lan-xu-2018-neural}. Textual entailment and natural language inference, in turn, consider both compatible and contradictory relations between given text pairs~\citep{dagan2005pascal,bowman-etal-2015-large,williams-etal-2018-broad,nie-etal-2020-adversarial}. Work on novelty and redundancy detection addresses a closely related notion of unique information by identifying content not present in previously observed text~\citep{schiffman-mckeown-2005-context,ghosal-etal-2018-novelty,ghosal-etal-2022-novelty}. More recently, researchers have examined how well LLMs handle these relations in longer and multi-source contexts, particularly when relevant evidence is distributed or conflicting~\citep{jiayang-etal-2024-econ,wan-etal-2024-evidence,kurfali-2025-conflicting}. However, these problems are studied separately, with the relevant text pair, claim, query, or evidence already provided. In contrast, we formulate these relations as a cross-narrative extraction problem and study whether LLMs can recover all shared, conflicting, and unique information directly from complete narratives.

\section{OUC Clause Extraction}
\label{section:task}
We introduce \textbf{Overlap--Unique--Conflict (OUC) extraction}, a novel task for multi-perspective narrative understanding. Let $N_A=\{S^A_1,\ldots,S^A_{|A|}\}$ and $N_B=\{S^B_1,\ldots,S^B_{|B|}\}$ denote the two narratives describing the same event or topic, where each $S^A_i$ and $S^B_j$ is an extractable textual span, corresponding to a sentence or clause from the source narrative. Given $N_A$ and $N_B$, the goal is to extract sets of verbatim clauses that capture information shared across the two narratives, information that conflicts between them, and information that appears in only one narrative. Specifically, we formulate the OUC extraction as:
\[
\mathcal{F}_{\mathrm{OUC}}(N_A,N_B)
=
\{O,C,U_A,U_B\},
\]
where the output consists of four sets: $O$ and $C$ contain overlapping and conflicting clause pairs, respectively, while $U_A$ and $U_B$ contain clauses unique to $N_A$ and $N_B$, respectively.

\subsection{Task Definitions}

\noindent\textbf{Overlap.}
A pair $(S^A_i,S^B_j)$ is considered overlapping when the two sentences/clauses refer to the same underlying event, fact, or aspect and convey mutually compatible information. The clauses need not use the same wording or provide the same level of detail; one may be more specific than the other as long as the additional information does not alter or contradict the shared content. Example: \textit{``The bill passed the Senate with bipartisan support'' and ``The Senate approved the bill with votes from both parties''} is an overlap pair because they express the same fact using different wording.

\noindent\textbf{Conflict.}
A pair $(S^A_i,S^B_j)$ is considered conflicting when the two clauses refer to the same underlying event, fact, or aspect but make mutually incompatible or contradictory claims about it. Differences in wording, emphasis, tone, or level of detail alone are not sufficient to constitute a conflict. For example, \textit{``The administration said the policy would reduce household costs'' and ``The administration acknowledged that the policy could increase costs for some households''} form a conflict pair because they make incompatible claims about the policy's economic effect.

For both overlap and conflict, alignments may be many-to-many: a sentence in one narrative can correspond to multiple sentences in the other, as long as each pairing independently satisfies the relevant relation definition. 

\noindent\textbf{Unique.}
A clause $S^A_i$ or $S^B_j$ is considered unique when the other narrative contains no clause that expresses corresponding information about the same underlying event, fact, or aspect. In other words, the clause cannot be paired with any clause in the other narrative as either an overlap or a conflict. We distinguish between \textbf{Unique-A}, for clauses that appear only in $N_A$, and \textbf{Unique-B}, for clauses that appear only in $N_B$.

\section{Benchmark Construction}
To our knowledge, no existing benchmark directly supports OUC clause extraction across alternative narratives. Constructing such a resource manually at scale is challenging and costly, as each narrative pair may contain multiple valid relations. 
Recent work, however, has shown that LLMs can be reliably used to generate high-quality synthetic data~\citep{gptJudge1,gptJudge2,gptJudge3,gptJudge4,gptJudge5}. Motivated by this, we construct a large-scale OUC benchmark using an LLM-assisted automated pipeline as a scalable alternative to human annotation.

\subsection{Data Source Selection}
To curate the benchmark, we sample narrative pairs from existing multi-document datasets, which are well-suited to our task because narratives about the same topic or event often contain shared, conflicting, and unique information. Specifically, we consider narratives from three domains: peer reviews, factual news, and political news. For the peer domain, we draw one narrative pair per paper from PeerSum~\citep{peersum}. For each paper, we select two reviews with different recommendation outcomes, such as \textit{Accepted--Borderline Accept} or \textit{Accepted--Rejected}. For factual news, we use WCEP~\citep{wcep} and select one article pair from each event instance in the disaster, accident, conflict, and attack categories. In both cases, we retain narrative pairs with semantic similarity scores between 0.5 and 0.8 to remove pairs that are either nearly identical or only weakly related. This yields 11,000 peer-review pairs and 8,000 factual-news pairs. To broaden the benchmark with political narratives, we additionally use AllSides~\citep{allside}, MultiOpEd~\citep{multioped}, and the politics subset of WCEP. These datasets provide 3,133, 1,192, and 1,975 pairs, respectively. Because the political datasets are small, we keep all available pairs rather than applying the same similarity filter. Overall, the initial pool contains approximately 25.3K narrative pairs.

\subsection{Automatic OUC Clause Curation}
We curate the OUC relation sets using an automated, multi-stage pipeline, as outlined below.

\noindent\textbf{Stage 1: Overlap and Conflict Candidate Extraction.}
We use \texttt{GPT-4.1-mini} as the primary extractor for identifying candidate overlap and conflict pairs. To reduce the chance of missing valid relations, we also employ \texttt{Mistral-Medium-3.5-128B} as a second extractor. Both models are independently queried with task-specific prompts to extract overlap and conflict pairs within the same narrative pair. We then take the union of the two models' outputs separately for overlap and conflict, yielding broader candidate lists for each relation. Afterward, we remove near-duplicate candidates within each relation using ROUGE-L~\citep{lin2004rouge}.  Two candidate pairs are considered near-duplicates when the corresponding clauses on both sides have ROUGE-L scores above 0.8. This stage intentionally prioritizes coverage, since false candidates can be filtered out during subsequent validation, whereas relations missed during candidate generation cannot be recovered.

\noindent\textbf{Stage 2: Candidate Validation.}
The increased coverage of the candidate pool comes at the cost of potentially incorrect alignments or relation assignments. We therefore subject every candidate to an independent validation step using \texttt{GPT-4.1-mini}, \texttt{Mistral-Medium-3.5-128B}, and \texttt{LLaMA-3.3-70B}. Given the source narratives and a candidate clause pair, each validator determines whether the candidate is valid for the relation under which it was extracted. We then aggregate the three judgments by majority vote and retain only candidates that are confirmed as valid.

\noindent\textbf{Stage 3: Unique-Clause Identification.}
Unique information is treated differently because, by definition, it lacks a corresponding clause in the other narrative. We therefore identify Unique-A and Unique-B only after the overlap and conflict relations have been validated. For this stage, we use \texttt{GPT-4.1-mini} with separate task-specific prompts to extract sets of Unique-A and Unique-B. In each case, the model receives both narratives together with the validated overlap and conflict pairs. 


\noindent\textbf{Stage 4: Final Consistency Check.}
Although the validated overlap and conflict pairs are provided during unique extraction, we perform an additional consistency check before finalizing the annotations. We compare each extracted unique clause against the clauses appearing in the validated overlap and conflict relations using ROUGE-L, and remove it when the similarity exceeds the same threshold of 0.8. This step serves as a final safeguard against residual cross-category assignments. The associated prompt templates are provided in Appendix~\ref{app:prompt-data-generation}. 

After applying the automatic pipeline, the initial pool of approximately 25.3K narrative pairs is reduced to 22,558, primarily because we exclude pairs for which the validated overlap set, conflict set, or both are empty. 
The remaining pairs and their curated overlap, conflict, Unique-A, and Unique-B relation sets constitute our silver-standard OUC benchmark. On average, each narrative pair contains 7.97 Overlap pairs, 2.17 Conflict pairs, 19.96 Unique-A clauses, and 18.61 Unique-B clauses. Detailed statistics are presented in Appendix Table~\ref{tab:dataset_distribution}.

\subsubsection{Human Validation}
To assess the quality of the silver-standard OUC benchmark, we conduct a human validation on a stratified sample of 120 narrative pairs, with 40 pairs drawn from each of the three domains. The sampled pairs contain a total of 5,388 automatically curated OUC instances, including 1,327 overlap, 695 conflict, 1,518 Unique-A, and 1,848 Unique-B instances. Three annotators voluntarily participate in the validation process. Before annotation, they were provided with detailed task guidelines (see Appendix~\ref{app:guidelines}) that defined each OUC relation and its corresponding decision criteria.  We measure inter-annotator agreement using Krippendorff's $\alpha$ and additionally compare human judgments with the automatically curated instances using exact-match agreement. Overall, agreement is high across all four relations: Krippendorff's $\alpha$ ranges from 0.880 to 0.948, while human agreement with the automatically curated instances ranges from 83.2\% to 96.2\%. Conflict shows the lowest agreement under both measures, indicating that validating conflicting information is comparatively more difficult than validating Overlap or Unique information. Full results are reported in Appendix Table~\ref{tab:human_eval}. 

While this study evaluates the validity of individual automatically curated instances, we further examine whether the curation pipeline recovers the complete set of OUC relations present in the narratives. For this purpose, we use a separate subset of 100 narrative pairs and ask one annotator to extract the complete set of OUC relations from the narrative pairs, which we then compare with the automatically curated sets using exact matching.  We achieve precision and recall above 90\% for all four relations, further supporting the benchmark's reliability. More details on validation and some examples from the dataset are provided in Appendix ~\ref{app:human_extraction} and ~\ref{app:data-samples}.



\section{Experiments}

\subsection{Benchmarking with Open-Source LLMs}

\noindent\textbf{Large Language Models.}
We evaluate a diverse set of 14 open-source LLMs spanning six prominent model families: LLaMA~\citep{dubey2024llama}, Phi~\citep{abdin2024phi}, OLMo~\citep{olmo2025olmo}, Qwen~\citep{yang2025qwen3}, Gemma~\citep{team2026gemma}, and Nemotron~\citep{blakeman2025nvidia}. The selected models cover a broad range of parameter scales, allowing us to examine OUC extraction capability across both model families and capacities. Specifically, we include LLaMA-3.2 (3B); Phi-4 (4B, 14B); OLMo-3 (7B) and OLMo-3.1 (32B); Qwen-3 (0.6B, 4B, 8B, 32B) and Qwen-3.6 (35B); Gemma-4 (2B, 4B, 31B); and Nemotron-3 (30B). For inference, we use greedy decoding across all models, with a temperature of 0 and a repetition penalty of 1.05.

\noindent\textbf{Methods.}
We compare three prompting-based methods: \textit{zero-shot}, \textit{few-shot}, and \textit{chain-of-thought (CoT)}. For each method, we extract Overlap, Conflict, Unique-A, and Unique-B separately using relation-specific instructions. We also consider a joint setting in which a single instruction extracts all OUC clauses simultaneously. 
Before applying these methods at scale, we conduct a pilot study on a small held-out subset of 60 narrative pairs to refine the instructions and select the most stable variant based on output consistency. During this study, we observed that native reasoning modes often produce excessively long reasoning traces without improving the quality of extraction. We therefore disable native reasoning for models that support it during the full evaluation. 

\noindent\textbf{Evaluation Protocol.}
We partition the OUC benchmark into training, validation, and test sets using a 75/5/20 split, yielding 16,916, 1,129, and 4,513 narrative pairs, respectively. We use the held-out test set exclusively for model evaluation. To measure OUC extraction performance, we compute macro Precision, Recall, and F1, with F1 as the primary metric. Because these metrics depend on matching predicted and gold extractions, we use relaxed matching with ROUGE-L, in which a prediction is considered correct if its ROUGE-L score against the corresponding gold item exceeds 0.6. For more details on metrics computation and threshold sensitivity, see Appendix~\ref{app:evaluation_metrics}.

\subsection{Learning OUC Extraction}
Beyond prompting, we investigate whether OUC extraction can be learned through task-specific supervision. To this end, we consider three post-training methods: direct supervision, preference learning, and reward-based optimization. All implementation details are provided in Appendix~\ref{app:implementation}.

\noindent\textbf{Supervised Fine-Tuning.} We perform supervised fine-tuning (SFT) for each OUC task separately. In every training instance, the input consists of the two source narratives and the corresponding task-specific instruction, while the target is the gold extraction for that task in JSON format. This directly trains the model to extract the complete set of task-specific clauses from the narrative pairs.

\noindent\textbf{Preference Learning.} We further employ task-specific direct preference optimization (DPO) \citep{rafailov2023direct}, which learns from relative preferences between alternative extractions rather than from a single target response. To construct this preference signal, we use the gold extraction as the preferred response and the predictions from a distractor model (LLaMA-3.2-1B) as the rejected response.  When no suitable model prediction is available, we instead generate a synthetic rejected response that is incorrect in content but similar in length and structure to the gold response. The full preference data construction procedure is provided in Appendix~\ref{app:dpo_construction}.


\noindent\textbf{Reward-Based Optimization.}
We also examine whether direct feedback on extraction quality can improve OUC extraction. Because each generated relation set can be compared directly with its corresponding gold set, the task provides an automatically verifiable reward signal and can therefore be naturally formulated within the reinforcement learning with verifiable rewards (RLVR) framework \citep{rlvr1,rlvr2,rlvr3,rlvr4,rlvr5}. For our task, we use the F1 score as the correctness reward because it jointly captures both missed and spurious extractions. We combine this with a binary reward for following the required output format. The final reward becomes $R = 0.9*R_{\text{correctness}} + 0.1*R_{\text{format}}$. To optimize the model policy with this reward, we use the Group Relative Policy Optimization (GRPO) algorithm \citep{shao2024deepseekmath}.



\section{Results and Analysis}
We organize our results and analysis around the following research questions: \textbf{RQ1) OUC Extraction Capability:} How well do current open-source LLMs perform on OUC extraction, and which OUC relation is the most challenging? \textbf{RQ2) Extraction Bottleneck:} What primarily limits Overlap and Conflict extraction performance? \textbf{RQ3) Task-Specific Post-Training:} Can task-specific learning improve OUC extraction in smaller models, and which supervision signal is most effective? \textbf{RQ4) Supervision Scaling:} Does scaling the amount of supervision affect OUC extraction performance?


\subsection{OUC Extraction Capability}
\noindent\textbf{Conflict is the most challenging extraction task.}
Table~\ref{tab:overall} shows a clear gap across the four relations: Unique-A and Unique-B are consistently easier, followed by Overlap, while Conflict remains the most difficult. This pattern persists even for larger models. For instance, under zero-shot prompting, Gemma-4-31B achieves F1 scores of 75.55\% and 76.75\% on the two Unique relations, compared with 61.13\% on Overlap and 48.58\% on Conflict.  Moreover, the difficulty is not solely due to missed clauses or pairs. Models often recover relevant pairs but also produce many incorrect ones, a pattern more pronounced in the Conflict task. 

\noindent\textbf{Prompting effects vary across models and tasks.}
No prompting strategy consistently improves performance across all models or all tasks. CoT yields notable gains on Overlap and Conflict for several models (e.g., Gemma-4-4B, Phi-4-14B, OLMo-3.1-32B, Qwen-3-32B, Qwen-3.6-35B, and Nemotron-3-30B), but the trend is not universal. One possible explanation is that CoT encourages the model to compare candidate clauses more explicitly before deciding whether they form an Overlap or Conflict pair. However, CoT often hurts the Unique extraction, where exhaustive coverage is more important. 
Few-shot prompting also fails to provide a consistent advantage across OUC extraction tasks, despite achieving the best individual F1 score of 61.29\% on the Overlap task. Consequently, zero-shot prompting remains the strongest overall choice on average across the four extraction tasks.

\noindent\textbf{Scaling helps, but family effects remain strong.}
Smaller models (i.e., 0.6B, 3B, 8B) are generally weaker on OUC extraction, particularly for Overlap and Conflict. This pattern becomes clearer when comparing different sizes within the same family. For instance, Qwen improves steadily from 0.6B to 35B in F1 scores, especially for the Conflict (from 1.23\% to 37.05\%) and Unique-A (from 14.20\% to 72.09\%) tasks under zero-shot prompting. Phi shows the same tendency from 4B to 14B. Even so, scale does not fully explain the results. For instance, OLMo-3.1-32B and Nemotron-3-30B remain relatively weak despite their size, whereas Gemma-4-31B achieves the strongest overall performance. This indicates that larger models generally help, but the benefit depends strongly on the model family.


\begin{table*}[!htb]
\centering
\scriptsize
\setlength{\tabcolsep}{5.5pt}

\caption{OUC extraction performance across prompting strategies. For each task, the highest and lowest F1-scores across the methods are highlighted in blue and orange, respectively. Due to space constraints, full precision, recall, and F1 results are reported in Appendix Table~\ref{tab:overall_full}.}
\label{tab:overall}

\begin{tabular}{l|cccc|cccc|cccc}
\toprule

\multirow{2}{*}{\textbf{LLM}}
& \multicolumn{4}{c|}{\textbf{Zero-shot}}
& \multicolumn{4}{c|}{\textbf{Few-shot}}
& \multicolumn{4}{c}{\textbf{Chain-of-Thought}} \\

\cmidrule(lr){2-5}
\cmidrule(lr){6-9}
\cmidrule(lr){10-13}

& \textbf{O} & \textbf{C} & $\mathbf{U_A}$ & $\mathbf{U_B}$
& \textbf{O} & \textbf{C} & $\mathbf{U_A}$ & $\mathbf{U_B}$
& \textbf{O} & \textbf{C} & $\mathbf{U_A}$ & $\mathbf{U_B}$ \\

\midrule

LLaMA-3.2-3B
& 18.34 & 7.54 & 36.29 & 28.26
& 15.12 & 6.19 & 33.64 & 28.08
& 17.69 & 7.86 & 27.20 & 27.70 \\

Phi-4-4B
& 11.06 & 5.90 & 36.26 & 38.17
& 13.95 & 3.68 & 38.54 & 14.94
& 17.52 & 6.22 & 32.99 & 29.02 \\

Phi-4-14B
& 44.61 & 30.33 & 69.72 & 71.64
& 42.24 & 28.35 & 64.28 & 62.63
& 49.37 & 29.69 & 59.73 & 67.79 \\

OLMo-3-7B
& 15.46 & 7.42 & 8.20 & 8.56
& 14.74 & 6.27 & 6.93 & \worstone{3.25}
& 13.45 & 9.76 & 11.90 & 8.27 \\

OLMo-3.1-32B
& 33.29 & 16.10 & 47.96 & 53.84
& 30.07 & 5.88 & 42.69 & 45.31
& 47.39 & 35.52 & 53.40 & 55.57 \\

Qwen-3-0.6B
& \worstone{3.04} & 1.23 & 14.20 & 4.75
& 4.09 & 1.22 & 11.59 & 5.66
& 4.80 & \worstone{0.08} & \worstone{5.21} & 4.00 \\

Qwen-3-4B
& 36.71 & 18.17 & 45.95 & 52.97
& 33.02 & 16.97 & 42.61 & 48.54
& 35.89 & 20.66 & 39.29 & 45.29 \\

Qwen-3-8B
& 40.71 & 18.53 & 63.66 & 64.48
& 38.78 & 18.09 & 50.91 & 46.55
& 37.14 & 22.72 & 22.66 & 30.01 \\

Qwen-3-32B
& 46.77 & 31.57 & 62.39 & 65.79
& 46.99 & 31.66 & 50.39 & 52.28
& 51.32 & 44.11 & 48.34 & 57.14 \\

Qwen-3.6-35B
& 53.42 & 37.05 & 72.09 & 74.54
& 53.28 & 39.01 & 69.22 & 70.10
& 59.44 & 45.92 & 62.56 & 49.13 \\

Gemma-4-2B
& 39.72 & 20.90 & 54.65 & 11.59
& 36.38 & 18.22 & 14.78 & 19.26
& 39.72 & 28.84 & 53.02 & 26.22 \\

Gemma-4-4B
& 46.41 & 28.22 & 68.43 & 61.66
& 44.48 & 28.58 & 65.19 & 59.54
& 54.05 & 39.47 & 66.36 & 68.74 \\

Gemma-4-31B
& 61.13 & \globalbest{48.58} & \globalbest{75.55} & \globalbest{76.75}
& \globalbest{61.29} & 48.27 & 73.39 & 73.64
& 60.19 & 46.12 & 62.76 & 66.98 \\

Nemotron-3-30B
& 12.73 & 5.11 & 45.46 & 41.24
& 12.43 & 5.19 & 39.79 & 35.49
& 36.50 & 14.71 & 28.59 & 33.61 \\

\bottomrule

\end{tabular}
\end{table*}


\subsection{Extraction Bottleneck}
To identify why Overlap and Conflict extraction is harder, we conduct two controlled diagnostic experiments on five representative models spanning different families and scales. We first isolate \textbf{\textit{relation understanding}} by giving the model a clause pair and asking, separately for Overlap and Conflict, whether the pair expresses the target relation or is invalid. This removes the need to search for relevant clauses or determine their correspondence. We then isolate \textbf{\textit{pair alignment}} by providing the set of ground-truth candidate clauses from each narrative and asking the model to identify which cross-narrative clause pairs form valid Overlap or Conflict relations. This setting will test only whether the models can construct the correct pairings among the given relevant clauses. We compare both diagnostics with the zero-shot end-to-end extraction performance reported in Table~\ref{tab:bottleneck}. 


\begin{table}[!htbp]
\centering
\scriptsize
\setlength{\tabcolsep}{3.5pt}

\caption{Bottleneck analysis for overlap and conflict extraction. Darker red shading indicates lower performance (macro F1-score), while blue indicates improvement.}
\label{tab:bottleneck}

\begin{tabular}{l|cc|cc|cc}
\toprule

\multirow{2}{*}{\textbf{LLM}}
& \multicolumn{2}{c|}{\textbf{Relation}}
& \multicolumn{2}{c|}{\textbf{Alignment}}
& \multicolumn{2}{c}{\textbf{Extraction}} \\

\cmidrule(lr){2-3}
\cmidrule(lr){4-5}
\cmidrule(lr){6-7}

& \textbf{O} & \textbf{C}
& \textbf{O} & \textbf{C}
& \textbf{O} & \textbf{C} \\

\midrule

Qwen-3-0.6B
& 76.35 & 58.91
& \scorecell{9.40} & \scorecell{45.25}
& \scorecell{3.04} & \scorecell{1.23} \\

LLaMA-3.2-3B
& 78.62 & 55.75
& \scorecell{43.98} & \gaincell{59.00}
& \scorecell{18.34} & \scorecell{7.54} \\

Qwen-3-8B
& 86.57 & 75.07
& \scorecell{62.21} & \scorecell{46.52}
& \scorecell{40.71} & \scorecell{18.53} \\

Phi-4-14B
& 89.30 & 86.52
& \scorecell{71.08} & \scorecell{79.38}
& \scorecell{44.61} & \scorecell{30.33} \\

Gemma-4-31B
& 89.80 & 87.28
& \scorecell{67.84} & \scorecell{74.93}
& \scorecell{61.13} & \scorecell{48.58} \\

\bottomrule
\end{tabular}

\end{table}

We observe that models are much better at recognizing Overlap and Conflict when a clause pair is already provided. Performance begins to drop once the model has to determine which clauses from the two narratives should be paired to form a valid Overlap or Conflict pair. For example, Qwen-3-8B drops $\sim$24\% (from 86.57\% to 62.21\% F1 score) on Overlap when moving from relation understanding to pair alignment. Interestingly, the drop is less consistent for Conflict. For example, Phi-4-14B and Gemma-4-31B retain relatively high alignment performance, while LLaMA-3.2-3B even improves slightly. One possible reason is that the conflict task often involves fewer plausible pairings, making the correspondence easier once the candidate clauses are given. In full extraction, however, performance drops substantially for both relations, suggesting that the main bottleneck lies in jointly discovering and aligning the relevant relation-specific information across narratives.

\paragraph{Retrieval-based candidate discovery.}
The alignment experiment above assumes access to gold candidate clauses and, therefore, removes the need to discover relevant information from the full narratives. To probe this step directly, we retrieve the top-$k$ sentences from the opposite narrative for each sentence using Qwen3-Embedding-0.6B and measure whether the gold counterpart is recovered. Recall increases with $k$: for Overlap, Recall@1 and Recall@5 are 63.49\% and 92.48\%, respectively, whereas for Conflict, they are 37.04\% and 79.97\%, respectively. Thus, retrieving only the nearest counterpart misses many gold relations, particularly for Conflict. Increasing $k$ improves coverage, but it also turns the task into candidate generation followed by additional pairwise filtering and relation classification, rather than directly probing whether an LLM can extract them from narratives.

\begin{table*}[!hb]
\centering
\scriptsize
\setlength{\tabcolsep}{4pt}

\caption{
OUC clause extraction performance across different post-training strategies. Here, values in parentheses denote the absolute change in macro F1-score relative to the best prompting baseline for each model and task. Blue and red indicate improvements and degradations, respectively. Detailed performance, including precision and recall scores, is provided in Appendix Table~\ref{tab:overall-posttraining}. 
}
\label{tab:posttraining-short}

\begin{tabular}{lcccc|cccc}
\toprule
\multirow{2}{*}{\textbf{Method}}
& \multicolumn{4}{c|}{\textbf{LLaMA-3.2-3B}}
& \multicolumn{4}{c}{\textbf{Qwen-3-0.6B}} \\

\cmidrule(lr){2-5}
\cmidrule(lr){6-9}

& \textbf{O} & \textbf{C} & $\mathbf{U_A}$ & $\mathbf{U_B}$
& \textbf{O} & \textbf{C} & $\mathbf{U_A}$ & $\mathbf{U_B}$ \\
\midrule

Baseline
& 18.34 & 7.86 & 36.29 & 28.26
& 4.80 & 1.23 & 14.20 & 5.66 \\

SFT
& 55.51\imp{37.17} & 42.19\imp{34.33} & 77.15\imp{40.86} & 78.92\imp{50.66}
& 50.20\imp{\textbf{45.40}} & 34.04\imp{\textbf{32.81}} & 75.73\imp{\textbf{61.53}} & 78.49\imp{\textbf{72.83}} \\

DPO
& 24.00\imp{5.66} & 6.93\dec{0.93} & 47.88\imp{11.59} & 45.59\imp{17.33}
& 18.72\imp{13.92} & 2.88\imp{1.65} & 7.97\dec{6.23} & 10.53\imp{4.87} \\

GRPO
& 28.74\imp{10.40} & 21.19\imp{13.33} & 69.95\imp{33.66} & 72.40\imp{44.14}
& 25.36\imp{20.56} & 5.52\imp{4.29} & 68.76\imp{54.56} & 67.52\imp{61.86} \\

\midrule
\addlinespace[2pt]

\multirow{2}{*}{}
& \multicolumn{4}{c|}{\textbf{Qwen-3-4B}}
& \multicolumn{4}{c}{\textbf{Qwen-3-8B}} \\

\cmidrule(lr){1-5}
\cmidrule(lr){6-9}

& \textbf{O} & \textbf{C} & $\mathbf{U_A}$ & $\mathbf{U_B}$
& \textbf{O} & \textbf{C} & $\mathbf{U_A}$ & $\mathbf{U_B}$ \\
\midrule

Baseline
& 36.71 & 20.66 & 45.95 & 52.97
& 40.71 & 22.72 & 63.66 & 64.48 \\

SFT
& 57.68\imp{20.97} & 48.37\imp{27.71} & 78.36\imp{32.41} & 80.87\imp{27.90}
& \textbf{59.98}\imp{19.27}
& \textbf{50.82}\imp{28.10}
& \textbf{79.31}\imp{15.65}
& \textbf{81.49}\imp{17.01} \\

DPO
& 42.81\imp{6.10} & 15.15\dec{5.51} & 60.80\imp{14.85} & 64.18\imp{11.21}
& 46.34\imp{5.63} & 18.62\dec{4.10} & 65.84\imp{2.18} & 67.18\imp{2.70} \\

GRPO
& 50.74\imp{14.03} & 33.89\imp{13.23} & 72.92\imp{26.97} & 75.65\imp{22.68}
& 55.58\imp{14.87} & 39.45\imp{16.73} & 74.60\imp{10.94} & 77.47\imp{12.99} \\

\bottomrule
\end{tabular}
\vspace{-3mm}
\end{table*}

\subsection{Task-Specific Post-Training}
To keep this study focused on models that are practical to fine-tune, we use LLaMA-3.2-3B and Qwen-3 at 0.6B, 4B, and 8B. These model sizes and families are also commonly used in prior work for parameter-efficient fine-tuning~\citep{zhang-etal-2026-skill,li-etal-2025-aide}. Table~\ref{tab:posttraining-short} reports OUC extraction performance under SFT, DPO, and GRPO, together with each model’s best prompting baseline.


\textbf{SFT shows the strongest and most consistent gains across model sizes.} We observe some of the largest gains for Qwen-3-0.6B, which performs poorly without task-specific training but improves by 32.81--72.83\% absolute points in F1 score across the four extraction tasks after SFT. These gains persist as model size increases, with Qwen-3-8B achieving the strongest results overall. Additionally, the fine-tuned Qwen-3-8B exceeds the best-performing model, Gemma-4-31B, on Conflict, Unique-A, and Unique-B, while remaining close on Overlap (59.98\% vs. 61.29\% F1 score). It also outperforms larger models, including Qwen-3-32B, Qwen-3.6-35B, OLMo-3.1-32B, and Nemotron-3-30B, across multiple tasks. These results suggest that task-specific direct supervision not only substantially improves OUC extraction but also enables smaller models to compete with much larger ones.

\textbf{GRPO improves OUC extraction, but its gains remain consistently below those of SFT.} Across models and tasks, GRPO yields clear improvements in F1-score over baselines. However, these gains are generally smaller than those obtained with supervised fine-tuning. One likely reason is that GRPO relies on a sequence-level reward largely based on overall F1, which captures the quality of the full extraction but provides limited guidance on which individual clauses or pairs to correct. \textbf{DPO, on the other hand, is considerably less stable.} While it improves some Overlap and Unique results, Conflict performance frequently drops below the prompting baseline. The possible reason is that DPO learns from response-level preference pairs that may differ in only a small number of extracted items, which could provide a weaker signal for correcting such localized errors. 

We additionally evaluate a joint OUC setting, in which all four tasks are predicted jointly rather than independently. Details are reported in Appendix~\ref{app:joint-ouc}.

\subsection{Supervision Scaling}
\begin{figure}[!htbp]
    \centering
    \includegraphics[width=0.55\textwidth]{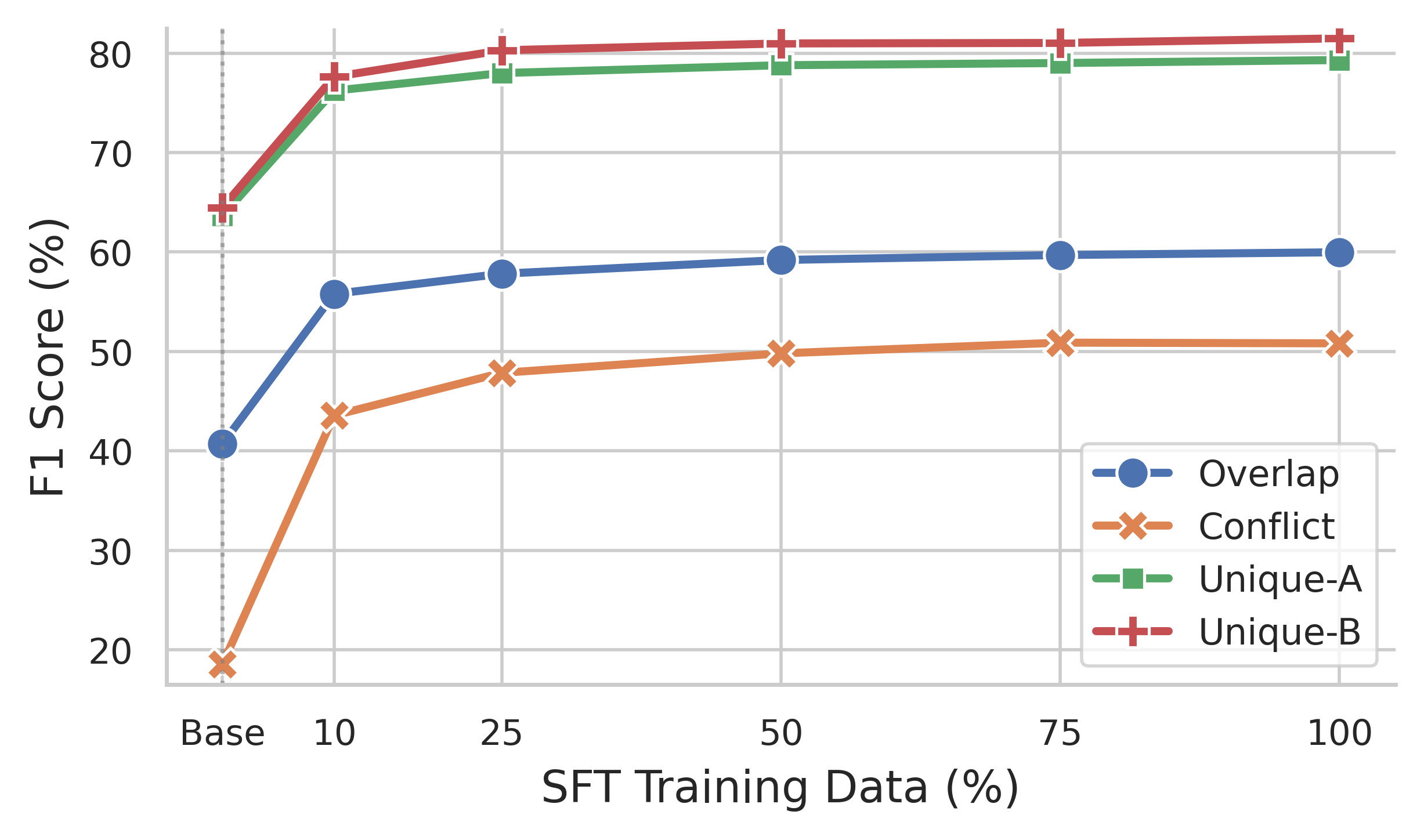}
    \caption{OUC extraction performance when the Qwen-3-8B model is fine-tuned across varying supervision sizes. The base represents the corresponding model zero-shot baseline.}
    \label{fig:sft-scalling}
\end{figure}

Since SFT performs best among the post-training methods, we next examine how much supervision is actually needed to obtain these gains. We fine-tune the Qwen-3-8B model by varying the training data from 10\% to 100\%. As shown in Figure~\ref{fig:sft-scalling}, most of the improvement appears with relatively little supervision. Even training with 10\% of the data produces a large jump over the baseline, and by 25\% the model already reaches an F1 score of 57.82\% on Overlap, 47.85\% on Conflict, and around 78--80\% on the two Unique tasks. However, after this point, additional data brings only modest gains. 
Notably, with only 25\% of the supervision, Qwen-3-8B is already competitive with the much larger models (i.e., Qwen-3.6-35B). This suggests that the strong performance is not simply due to training on a large dataset; rather, a relatively small amount of data already yields most of the gains. We further examine the cross-domain generalization: details are provided in Appendix~\ref{app:cross-domain}.

\subsection{Error Analysis}
To better understand where the model struggles with OUC extraction, we conduct a qualitative error analysis. Specifically, we analyze 300 errors from Gemma-4-31B predictions, with 75 cases sampled from each task across 90 narrative pairs. The most frequent category we found is \textit{Span/Verbatim Errors} (22.7\%), where the model identifies relevant content but extracts an incorrect span, changes the level of detail, or produces a non-verbatim form. We also find \textit{Missed Information} and \textit{Wrong Semantic Match} in 21.7\% of the cases each. Missed Information reflects cases where a valid clause or pair is not extracted, while Wrong Semantic Match occurs when the model selects clauses that are topically or semantically related but do not satisfy the target Overlap or Conflict relation. Another recurring problem appears in the Unique tasks, where in 15.0\% of cases the model labels a clause as Unique even though related information is present elsewhere in the narrative. We refer to these as \textit{Missed Counterpart errors}. We further observe \textit{Wrong Pairing} in 13.3\% of the cases, where relevant information is present in both narratives, but the model links the wrong counterparts. \textit{Relation Confusion} accounts for 5.7\%, covering cases where the correct clause pair is identified but assigned the wrong relation, such as predicting Conflict for an Overlap pair. Across these categories, \textit{implicit semantic relations} appear in 23.3\% of the analyzed cases. In such examples, corresponding information is expressed with substantially different wording, making the connection difficult to identify from surface similarity alone. These cases often result in missed pairs, incorrect matches, or false Unique predictions. Some examples from these error categories are shown in Table~\ref{tab:error_examples}.
\section{Conclusion}
In this work, we introduced the Overlap--Unique--Conflict (OUC) clause extraction task and a new benchmark for extracting this information directly from alternative narratives. Across 14 open-source LLMs, our results show that current models, particularly smaller ones, remain much weaker at extracting Overlap and Conflict pairs from full narratives than at recognizing these relations when the relevant information is already given. Importantly, task-specific supervision substantially narrows this gap. A fine-tuned Qwen-3-8B improves F1 score by roughly 15--28\% across all tasks and can match or outperform models that are roughly four times larger, although the extraction gap is not fully eliminated. These findings point to cross-narrative information discovery and pairing as a key limitation of current LLMs and an important direction for future work.

\section*{Limitations}
Our current OUC formulation is limited to sentence-level relations between narrative pairs. It does not consider finer-grained claim-level extraction, where a sentence may need to be decomposed before determining whether the information is overlapping, conflicting, or unique. Also, it doesn't capture cases where relevant evidence spans multiple sentences or more than two narratives. Finally, our diagnostic experiments identify candidate discovery and cross-narrative alignment as important sources of difficulty. Although post-training improves performance, these bottlenecks remain, leaving room for methods that address them more explicitly.

\subsection*{AI Use Statement}
In this work, we used generative AI tools to improve the grammar, clarity, readability, and organization of the manuscript's human-written sections. The tools were primarily used to identify unclear passages, assess whether the intended meaning was conveyed, and suggest improvements to presentation and structure. They were not used to generate experimental results, annotations, or scientific claims. All AI-assisted revisions were reviewed and edited by the authors, who take full responsibility for the final content of the paper.



\subsection*{Reproducibility statement}
To support reproducibility, we provide detailed descriptions of the OUC task,
dataset construction, annotation procedure, and evaluation protocol in the main paper and appendix. The appendix further documents the annotation guidelines, implementation framework, training setup, hyperparameters, decoding settings, hardware configuration, and computational cost for SFT, DPO, and GRPO. We also report the prompts and task-specific training configurations used in our experiments. Upon publication, we plan to release the dataset, annotation resources, and code required to reproduce the reported experiments.



\bibliography{references}
\bibliographystyle{Style/iclr2027_conference}

\appendix
\section{Appendix}

\begin{table}[!htbp]
\centering
\small
\setlength{\tabcolsep}{5pt}

\caption{Distribution of narrative pairs and extracted OUC instances across the three domains.}
\label{tab:dataset_distribution}

\begin{tabular}{lcccccc}
\toprule
\textbf{Domain} &
\textbf{\# Pairs} &
\textbf{Avg. Sent./Doc.} &
\textbf{Overlap} &
\textbf{Conflict} &
\textbf{Unique-A} &
\textbf{Unique-B} \\
\midrule
Political & 5,665 & 29.8 & 44,685  & 10,713 & 113,383 & 100,224 \\
Factual   & 6,753 & 31.6 & 49,734 & 13,671 & 134,287 & 137,457 \\
Peer      & 10,140 & 34.7 & 85,371 & 24,473 & 202,679 & 182,222 \\
\midrule
Overall   & 22,558 & 32.7 & 179,790 & 48,857 & 450,349 & 419,903 \\
\bottomrule
\end{tabular}

\end{table}

\begin{table}[!htb]
\centering
\small
\setlength{\tabcolsep}{4pt}

\caption{Inter-annotator agreement and human agreement with automatically curated OUC instances across relations and domains. Here, $\alpha$ denotes Krippendorff's alpha among the human annotators, and H--Auto denotes the mean exact-match agreement between human judgments and the automatically curated instances. The standard deviation of H--Auto across annotators ranges from 0.000 to 0.012.}
\label{tab:human_eval}
\begin{tabular}{l|cc|cc|cc|cc}
\toprule
& \multicolumn{2}{c}{\textbf{Overlap}}
& \multicolumn{2}{c}{\textbf{Conflict}}
& \multicolumn{2}{c}{\textbf{Unique-A}}
& \multicolumn{2}{c}{\textbf{Unique-B}} \\
\cmidrule(lr){2-3}
\cmidrule(lr){4-5}
\cmidrule(lr){6-7}
\cmidrule(lr){8-9}

\textbf{Domain}
& $\alpha$ & H--Auto
& $\alpha$ & H--Auto
& $\alpha$ & H--Auto
& $\alpha$ & H--Auto \\
\midrule

Accident
& 0.939 & 0.904
& 0.836 & 0.776
& 0.918 & 0.975
& 0.804 & 0.971 \\

Peer
& 0.931 & 0.959
& 0.939 & 0.804
& 0.964 & 0.920
& 0.948 & 0.907 \\

Side
& 0.932 & 0.965
& 0.862 & 0.869
& 0.950 & 0.970
& 0.953 & 0.987 \\

\midrule
\rowcolor{gray!15}
\textbf{Overall}
& 0.934 & 0.946
& 0.880 & 0.832
& 0.948 & 0.954
& 0.932 & 0.962 \\

\bottomrule
\end{tabular}

\end{table}

\subsection{Validation with Human Extracted Subset}
\label{app:human_extraction}
Table~\ref{tab:full_human_validation} reports precision, recall, and F1 computed by comparing the human-extracted and automatically extracted OUC clauses from 100 narrative pairs, including 25 peer-review, 25 factual, and 50 political narrative pairs. We ensure that the human-annotated subset remains unseen during training and can be used to evaluate both prompted and fine-tuned models without train--test leakage. 
\begin{table}[!htb]
\centering
\small
\begin{tabular}{lrrrrrr}
\toprule
Relation & Auto & Human & Matched & Precision & Recall & F1 \\
\midrule
Overlap  & 562  & 559  & 544  & 96.80 & 97.32 & 97.06 \\
Conflict & 177  & 182  & 168  & 94.92 & 92.31 & 93.59 \\
Unique-A & 1468 & 1459 & 1452 & 98.91 & 99.52 & 99.21 \\
Unique-B & 1499 & 1492 & 1483 & 98.93 & 99.40 & 99.16 \\
\midrule
\textbf{Overall} & \textbf{3706} & \textbf{3692} & \textbf{3647} & \textbf{98.41} & \textbf{98.78} & \textbf{98.59} \\
\bottomrule
\end{tabular}
\caption{Comparison between automatically curated and human-extracted OUC instances on the fully annotated subset. Precision and recall are computed using an exact match.}
\label{tab:full_human_validation}
\end{table}

We observe consistently high agreement across all four relations, with F1 above 97\% for Overlap and above 99\% for both Unique relations. Conflict remains the most difficult relation, with an F1 score of 93.59, consistent with our instance-level validation results. Overall, the automatically curated annotations achieve 98.41\% precision, 98.78\% recall, and 98.59\% F1 against the human-extracted sets. A closer inspection of the remaining differences shows that human annotators occasionally miss valid OUC pairs entirely, particularly when the relation is subtle and requires repeatedly looking back and forth across the two narratives. These cases often involve information that is not lexically obvious and can only be identified by connecting context across multiple sentences. This suggests that exhaustive OUC extraction places a non-trivial cognitive burden on annotators, who must continuously search, align, and compare information across both narratives. Despite this difficulty, the close agreement between human extraction and the automatically curated sets provides further evidence that the curation pipeline captures the underlying OUC relations with high fidelity.

\subsection{Implementation and Training Details}
\label{app:implementation}
In our experiments, we use the \texttt{instruct} variants of all LLMs. Models and tokenizers are loaded with the Hugging Face \texttt{Transformers} library,
while inference is performed with vLLM~\citep{vllm} using a batch size of 32. We use greedy decoding by setting the temperature to 0 and keeping each model's default decoding configuration for the remaining generation parameters, including top-$p$ and repetition penalty. For post-training, we adopt parameter-efficient fine-tuning with LoRA. SFT and DPO are implemented with the TRL library using \texttt{SFTTrainer} and
\texttt{DPOTrainer}, respectively, whereas GRPO is implemented with EasyR1, which builds on the VeRL~\citep{verl} framework for distributed reinforcement learning. All training is conducted in BF16 precision on NVIDIA H100 80\,GB GPUs. SFT and DPO use a single H100 GPU per run, while GRPO uses four H100 GPUs with FSDP-based distributed training.  We use gradient checkpointing for both SFT and DPO to reduce memory usage during training. DPO, in particular, is optimized with the sigmoid preference loss using $\beta=0.1$. GRPO, on the other hand, relies on stochastic rollout generation, for which we use a sampling temperature of $1.0$. Across all post-training methods, checkpoints are evaluated and saved every 500 steps, and the checkpoint with the lowest validation loss is selected for final evaluation.  The main hyperparameters for all three
Post-training methods are summarized in Table~\ref{tab:training-hyperparameters}.

\begin{table}[!htbp]
\centering
\small
\setlength{\tabcolsep}{4.5pt}
\renewcommand{\arraystretch}{1.08}
\caption{
Main hyperparameters used for SFT, DPO, and GRPO.
$^\dagger$For GRPO, the two values denote the maximum prompt and response
lengths, respectively.
}
\label{tab:training-hyperparameters}

\begin{tabular}{lccc}
\toprule
\textbf{Hyperparameter} & \textbf{SFT} & \textbf{DPO} & \textbf{GRPO} \\
\midrule
Epochs & 3 & 3 & 3 \\
Learning rate & $2\times10^{-4}$ & $5\times10^{-6}$ & $5\times10^{-6}$ \\
Per-device batch size & 1 & 1 & -- \\
Gradient accumulation & 4 & 4 & -- \\
Effective / global batch size & 4 & 4 & 16 \\
LoRA rank ($r$) & 16 & 16 & 16 \\
LoRA $\alpha$ & 32 & 32 & 32 \\
LoRA dropout & 0.05 & 0.05 & -- \\
Maximum length & 9144 & 9144 & 6,144 / 3,000 \\
Warmup ratio & 0.05 & 0.05 & -- \\
Weight decay & 0.01 & 0.01 & 0.01 \\
DPO $\beta$ & -- & 0.1 & -- \\
Rollouts per prompt & -- & -- & 8 \\
KL coefficient & -- & -- & 0.01 \\
\bottomrule
\end{tabular}

\end{table}

\paragraph{Compute Cost.}
Training time varies substantially across post-training methods. Based on the observed per-task runtimes, a three-epoch training run is estimated to require
approximately 11.0 hours for SFT and 32.6 hours for DPO on a single H100 GPU. GRPO is considerably more expensive; three epochs require approximately 80.0 hours on average, using 2--4 H100 GPUs depending on the task. For inference, Overlap and Conflict typically require approximately 6--10 hours each, while Unique-A and Unique-B require approximately 10--12 hours each. Thus, evaluating all four OUC tasks for a model requires roughly 36--44 wall-clock hours on a single H100 GPU.

\subsection{Controlled DPO Preference Construction}
\label{app:dpo_construction}
For each OUC relation, we construct preference tuples $(x,y^{+},y^{-})$, where $x$ contains the relation-specific instruction and the two source narratives, $y^{+}$ is the gold extraction, and $y^{-}$ is an incorrect but structurally similar alternative. The preference data are designed so that the learning signal reflects extraction correctness rather than simple properties such as output length. For the rejected pair, we select it from the predictions of a distractor model (LLaMA-3.2-1B). We retain only predictions that contain a particular extraction error. When no suitable model-generated prediction is available, we construct a minimally corrupted alternative using hard distractors from the same narrative pair. The detailed construction procedure is described below.


\paragraph{Model-Generated Negatives.}
We first obtain candidate-rejected responses from another weaker model (LLaMA-3.2-1B), and compare each prediction $P$ with the gold set $G$. Incorrect predictions are grouped into three error types: an \textit{omission}, where one or more gold items are missing without introducing incorrect items; a \textit{false positive}, where incorrect items are added; and a \textit{replacement}, where gold items are substituted with incorrect ones while keeping the output size approximately unchanged. Correct predictions, empty outputs, and predictions that differ substantially from the gold response in size are not used directly as rejected responses.

To keep model-generated negatives close to the preferred response, we constrain the allowable difference in the number of extracted items for omission and false-positive errors as
\[
\Delta_{\max}
=
\max\left(
1,
\operatorname{round}(0.25|G|)
\right),
\]
where $|G|$ denotes the number of gold items. Replacement errors naturally provide stronger control over output length because they preserve the overall number of extracted items.

\paragraph{Error-Type Distribution.}
Because the naturally occurring errors are uneven across relations, we control the proportion of omission, false-positive, and replacement negatives used for each OUC task. Replacement negatives receive the largest share because they make output length less informative for distinguishing preferred and rejected responses. Table~\ref{tab:dpo_error_distribution} reports the resulting target distribution.

\begin{table}[!htb]
\centering
\small

\caption{Target distribution of rejected-response error types used for DPO preference construction.}
\label{tab:dpo_error_distribution}
\begin{tabular}{lccc}
\toprule
\textbf{Relation} & \textbf{Replacement} & \textbf{Omission} & \textbf{False Positive} \\
\midrule
Overlap   & 50\% & 25\% & 25\% \\
Conflict  & 45\% & 15\% & 40\% \\
Unique-A  & 55\% & 30\% & 15\% \\
Unique-B  & 55\% & 30\% & 15\% \\
\bottomrule
\end{tabular}

\end{table}

\paragraph{Synthetic Negatives.}
When a suitable model-generated negative is unavailable, we construct a rejected response by minimally perturbing the gold extraction while preserving its overall structure. Depending on the target error type, we remove a small number of gold items, add incorrect items, or replace an equal number of gold items with incorrect ones. To keep the synthetic negatives plausible, we select incorrect items from the same narrative pair rather than from unrelated examples. We choose these hard distractors according to the target relation: for Conflict, validated overlap pairs provide related but non-conflicting alternatives; for Overlap, conflict pairs or misaligned clause combinations provide semantically related but invalid matches; and for Unique-A and Unique-B, clauses participating in validated overlap or conflict relations provide plausible but non-unique alternatives. If the assigned error type cannot be created, we instead construct a replacement negative. If no valid rejected response can be produced, we remove that instance from the DPO training set.

\subsection{Details on Evaluation Metrics}
\label{app:evaluation_metrics}

\subsubsection{Metrics Computation}

\begin{algorithm}[!b]
\caption{Macro Precision, Recall, and F1 for OUC Extraction}
\label{alg:macro-prf}
\small
\begin{algorithmic}[1]

\Require Samples $\mathcal{S}$, predictions $\hat{\mathcal{Y}}$, ground truth $\mathcal{Y}$,
similarity threshold $\tau$
\Require Categories
$\mathcal{C}=\{\textsc{Overlap},\textsc{Conflict},
\textsc{Unique-A},\textsc{Unique-B}\}$

\ForAll{$c \in \mathcal{C}$}
    \State $\mathcal{P},\mathcal{R},\mathcal{F} \gets [\,],[\,],[\,]$

    \ForAll{$s \in \mathcal{S}$}
        \State $\hat{Y} \gets$ predicted items for category $c$ in sample $s$
        \State $Y \gets$ gold items for category $c$ in sample $s$
        \State $E \gets [\,]$

        \ForAll{$(\hat{y}_i,y_j) \in \hat{Y}\times Y$}
            \State $\sigma_{ij} \gets
            \Call{Similarity}{\hat{y}_i,y_j,c}$
            \If{$\sigma_{ij} \ge \tau$}
                \State Add $(\sigma_{ij},i,j)$ to $E$
            \EndIf
        \EndFor

        \State Sort $E$ by decreasing $\sigma$
        \State $M \gets \emptyset$
        \State $I_{\text{pred}} \gets \emptyset$
        \State $I_{\text{gold}} \gets \emptyset$

        \ForAll{$(\sigma,i,j) \in E$}
            \If{$i \notin I_{\text{pred}}$ \textbf{and}
                $j \notin I_{\text{gold}}$}
                \State $M \gets M \cup \{(i,j)\}$
                \State $I_{\text{pred}} \gets I_{\text{pred}} \cup \{i\}$
                \State $I_{\text{gold}} \gets I_{\text{gold}} \cup \{j\}$
            \EndIf
        \EndFor

        \State $m \gets |M|$

        \State $P_s \gets
        \begin{cases}
        m/|\hat{Y}|, & |\hat{Y}|>0\\
        0, & \text{otherwise}
        \end{cases}$

        \State $R_s \gets
        \begin{cases}
        m/|Y|, & |Y|>0\\
        0, & \text{otherwise}
        \end{cases}$

        \State $F_s \gets
        \begin{cases}
        \frac{2P_sR_s}{P_s+R_s}, & P_s+R_s>0\\
        0, & \text{otherwise}
        \end{cases}$

        \State Append $P_s$, $R_s$, and $F_s$ to
        $\mathcal{P}$, $\mathcal{R}$, and $\mathcal{F}$

    \EndFor

    \State $\mathrm{MacroP}_{c}
    \gets \frac{1}{|\mathcal{S}|}\sum_{s \in \mathcal{S}} P_s$

    \State $\mathrm{MacroR}_{c}
    \gets \frac{1}{|\mathcal{S}|}\sum_{s \in \mathcal{S}} R_s$

    \State $\mathrm{MacroF1}_{c}
    \gets \frac{1}{|\mathcal{S}|}\sum_{s \in \mathcal{S}} F_s$

\EndFor

\end{algorithmic}
\end{algorithm}

We evaluate each OUC category independently at the sample level. For each sample, we compare every predicted item against every gold item using the category-specific similarity function. Prediction--gold pairs with similarity at least $\tau$ are retained as candidate matches. We then sort these candidates by similarity and greedily construct a one-to-one matching, such that each prediction and each gold item can participate in at most one match. Algorithm~\ref{alg:macro-prf} summarizes the full evaluation procedure.

Let $\mathcal{S}$ denote the set of evaluation samples. For each sample $s \in \mathcal{S}$, let $\hat{Y}_s$ and $Y_s$ denote the predicted and gold sets, respectively, and let $M_s$ denote the resulting set of matched prediction--gold pairs. We compute sample-level precision, recall, and F1 as

\[
P_s = \frac{|M_s|}{|\hat{Y}_s|},
\qquad
R_s = \frac{|M_s|}{|Y_s|},
\qquad
F1_s = \frac{2P_sR_s}{P_s+R_s}.
\]

If $|\hat{Y}_s|=0$, precision is set to zero; if $|Y_s|=0$, recall is set to zero. When $P_s+R_s=0$, $F1_s$ is also set to zero. We report macro scores by averaging the sample-level precision, recall, and F1 values across all evaluation samples:

\[
\mathrm{MacroP}
= \frac{1}{|\mathcal{S}|}\sum_{s \in \mathcal{S}} P_s,
\qquad
\mathrm{MacroR}
= \frac{1}{|\mathcal{S}|}\sum_{s \in \mathcal{S}} R_s,
\qquad
\mathrm{MacroF1}
= \frac{1}{|\mathcal{S}|}\sum_{s \in \mathcal{S}} F1_s.
\]

\noindent\textbf{Pair similarity.}
For Overlap and Conflict, let
$\hat{y}=(\hat{s}_1,\hat{s}_2)$ denote a predicted pair and $y=(s_1,s_2)$ denote a gold pair. We compute

\[
\operatorname{sim}(\hat{y},y)
=
\min\left(
\operatorname{ROUGE\text{-}L}_{F1}(\hat{s}_1,s_1),
\operatorname{ROUGE\text{-}L}_{F1}(\hat{s}_2,s_2)
\right).
\]

Thus, both sentences in a predicted pair must be sufficiently similar to their corresponding gold sentences. For Unique-A and Unique-B, similarity is computed only on the sentence from the corresponding narrative:

\[
\operatorname{sim}_{U_A}(\hat{y},y)
=
\operatorname{ROUGE\text{-}L}_{F1}(\hat{s}_1,s_1),
\qquad
\operatorname{sim}_{U_B}(\hat{y},y)
=
\operatorname{ROUGE\text{-}L}_{F1}(\hat{s}_2,s_2).
\]

\subsubsection{Threshold Sensitivity}
Table~\ref{tab:threshold-sensitivity} shows that threshold selection has a noticeable effect on the absolute F1 scores. As the ROUGE-L-based matching threshold increases, performance decreases consistently because predictions with partial lexical overlap to the reference are less likely to satisfy the matching criterion. The decrease is larger for Overlap and Conflict than for the Unique categories. Although the absolute scores vary with the threshold, the overall performance pattern remains consistent across settings, suggesting that the main conclusions are not dependent on a particular threshold choice.

\begin{table}[!ht]
\centering
\small
\setlength{\tabcolsep}{6pt}
\renewcommand{\arraystretch}{1.08}

\caption{Threshold sensitivity analysis on the best prompting-based model and the best post-trained model. All reported values are macro F1-scores.}
\label{tab:threshold-sensitivity}

\begin{tabular}{lc|cccc}
\toprule
\textbf{Model} & \textbf{Threshold}
& \textbf{Overlap} & \textbf{Conflict}
& $\mathbf{Unique_A}$ & $\mathbf{Unique_B}$ \\
\midrule

\multirow{3}{*}{Gemma-4-31B}
& 0.60 & 61.13 & 48.58 & 75.55 & 76.75 \\
& 0.75 & 58.27 & 46.53 & 74.22 & 75.44 \\
& 0.90 & 56.07 & 44.78 & 72.63 & 74.03 \\

\midrule

\multirow{3}{*}{Qwen-3-8B-SFT}
& 0.60 & 59.98 & 50.82 & 79.31 & 81.49 \\
& 0.75 & 56.08 & 48.33 & 77.69 & 79.97 \\
& 0.90 & 54.24 & 46.61 & 76.14 & 78.74 \\

\bottomrule
\end{tabular}
\end{table}

\subsection{OUC Extraction Performance Across Domains}
\label{app:domain-performance}


\begin{table}[!htb]
\centering
\tiny
\setlength{\tabcolsep}{3.5pt}
\renewcommand{\arraystretch}{1.05}

\caption{OUC extraction performance (macro F1-score) across three domains for five representative models under the two best prompting strategies. Blue and orange shading indicate gains and drops with CoT, respectively.}
\label{tab:cross-dataset-representative}

\begin{tabular}{l|cccc|cccc}
\toprule

\multirow{2}{*}{\textbf{LLM}}
& \multicolumn{4}{c|}{\textbf{Zero-shot}}
& \multicolumn{4}{c}{\textbf{CoT}} \\

\cmidrule(lr){2-5}
\cmidrule(lr){6-9}

& \textbf{O} & \textbf{C} & $\mathbf{U_A}$ & $\mathbf{U_B}$
& \textbf{O} & \textbf{C} & $\mathbf{U_A}$ & $\mathbf{U_B}$ \\


\midrule
\addlinespace[1pt]
\rowcolor{gray!20}
\multicolumn{9}{c}{\textbf{\textit{Peer}}} \\
\addlinespace[-0.5pt]
\midrule

Qwen-3-0.6B
& 2.43 & 0.58 & 15.25 & 6.47
& \impcell{4.18}{1.75}
& \deccell{0.16}{0.42}
& \deccell{6.24}{9.01}
& \deccell{4.82}{1.65} \\

LLaMA-3.2-3B
& 18.64 & 7.84 & 23.61 & 18.04
& \impcell{18.97}{0.33}
& \impcell{9.18}{1.34}
& \impcell{32.38}{8.77}
& \impcell{25.91}{7.87} \\

Qwen-3-8B
& 39.34 & 18.87 & 57.08 & 53.74
& \deccell{39.12}{0.22}
& \impcell{24.85}{5.98}
& \deccell{24.48}{32.60}
& \deccell{27.57}{26.17} \\

Phi-4-14B
& 42.36 & 32.09 & 65.29 & 63.53
& \impcell{46.50}{4.14}
& \deccell{26.82}{5.27}
& \deccell{53.19}{12.10}
& \deccell{61.40}{2.13} \\

Gemma-4-31B
& 59.50 & 48.44 & 71.66 & 69.89
& \deccell{59.14}{0.36}
& \deccell{47.42}{1.02}
& \deccell{51.88}{19.78}
& \deccell{55.05}{14.84} \\


\midrule
\addlinespace[1pt]
\rowcolor{gray!20}
\multicolumn{9}{c}{\textbf{\textit{Factual}}} \\
\addlinespace[-0.5pt]
\midrule

Qwen-3-0.6B
& 3.38 & 2.54 & 14.87 & 3.80
& \impcell{4.70}{1.32}
& \deccell{1.61}{0.93}
& \deccell{4.63}{10.24}
& \deccell{3.72}{0.08} \\

LLaMA-3.2-3B
& 17.85 & 6.63 & 49.13 & 34.72
& \deccell{15.96}{1.89}
& \impcell{6.89}{0.26}
& \deccell{24.03}{25.10}
& \deccell{28.08}{6.64} \\

Qwen-3-8B
& 39.77 & 19.13 & 71.49 & 72.16
& \deccell{33.05}{6.72}
& \impcell{21.96}{2.83}
& \deccell{22.29}{49.20}
& \deccell{29.97}{42.19} \\

Phi-4-14B
& 44.73 & 30.21 & 75.03 & 77.89
& \impcell{50.58}{5.85}
& \impcell{33.38}{3.17}
& \deccell{67.38}{7.65}
& \deccell{71.76}{6.13} \\

Gemma-4-31B
& 60.80 & 49.72 & 79.72 & 82.04
& \deccell{59.48}{1.32}
& \deccell{45.36}{4.36}
& \deccell{72.83}{6.89}
& \deccell{75.94}{6.10} \\


\midrule
\addlinespace[1pt]
\rowcolor{gray!20}
\multicolumn{9}{c}{\textbf{\textit{Political}}} \\
\addlinespace[-0.5pt]
\midrule

Qwen-3-0.6B
& 3.72 & 0.86 & 11.55 & 2.79
& \impcell{6.04}{2.32}
& \impcell{1.32}{0.46}
& \deccell{4.06}{7.49}
& \impcell{3.11}{0.32} \\

LLaMA-3.2-3B
& 18.40 & 8.10 & 43.72 & 38.89
& \deccell{17.03}{1.37}
& \deccell{6.66}{1.44}
& \deccell{21.76}{21.96}
& \deccell{30.41}{8.48} \\

Qwen-3-8B
& 44.28 & 17.20 & 66.13 & 74.57
& \deccell{38.44}{5.84}
& \impcell{19.82}{2.62}
& \deccell{19.84}{46.29}
& \deccell{34.42}{40.15} \\

Phi-4-14B
& 48.50 & 27.30 & 71.36 & 78.74
& \impcell{53.08}{4.58}
& \impcell{30.42}{3.12}
& \deccell{62.36}{9.00}
& \deccell{74.50}{4.24} \\

Gemma-4-31B
& 64.45 & 47.48 & 77.57 & 82.73
& \deccell{62.91}{1.54}
& \deccell{44.70}{2.78}
& \deccell{70.25}{7.32}
& \deccell{77.71}{5.02} \\

\bottomrule
\end{tabular}

\end{table}

Table~\ref{tab:cross-dataset-representative} reports OUC extraction performance across domains. A consistent pattern is that peer-review narratives are generally more challenging, particularly for Unique extraction. For example, under zero-shot prompting, Gemma-4-31B obtains 71.66\% and 69.89\% F1 on Unique-A and Unique-B in the peer domain, compared with 79.72\%/82.04\% on factual and 77.57\%/82.73\% on political narratives. Other models also show a similar gap. One possible reason is that peer reviews often contain semantically related criticisms that are not exact counterparts. For example, one review may criticize the paper's motivation, while another questions the clarity of the method. Although these comments are related, they should still be treated as distinct information. This makes it harder to determine whether a clause is truly unique to a single review. Factual and political narratives, in contrast, tend to contain more directly stated event details, such as what happened, who was involved, and specific outcomes, which can make corresponding information easier to locate across narratives. Even so, the relative difficulty of the OUC tasks remains similar across domains: Overlap is challenging, while Conflict is consistently the hardest. This suggests that identifying and pairing conflicting information remains difficult regardless of the type of narrative.

CoT has a mixed effect across the four OUC tasks. It can improve Overlap and Conflict for some models, especially Phi-4-14B, which gains 4.14--5.85\% F1 score on Overlap across the three domains and also improves Conflict in the factual and political domains. On the other hand, CoT often reduces performance on the Unique tasks. For Qwen-3-8B, for example, Unique-A drops by 32.60--49.20\% F1 score across the three domains. This suggests that the benefit of CoT is not uniform across OUC relations: it can help with some pairwise comparisons, but it can also make it harder to preserve information specific to one narrative. The same pattern is not tied to model size, as Gemma-4-31B does not improve with CoT on any of the four tasks across the three domains.

\begin{table}[!htb]
\centering
\scriptsize
\setlength{\tabcolsep}{5.5pt}

\caption{OUC extraction performance on the automatically curated and human-extracted gold sets. \textbf{Bold} and \underline{underlined} F1 values indicate the best and second-best settings, respectively, for each task within a model family.}
\begin{tabular}{lllccc|ccc}
\toprule
\multirow{2}{*}{Model} &
\multirow{2}{*}{Method} &
\multirow{2}{*}{Task} &
\multicolumn{3}{c|}{Automatic} &
\multicolumn{3}{c}{Human} \\
\cmidrule(lr){4-6}\cmidrule(lr){7-9}
& & & P & R & F1 & P & R & F1 \\
\midrule

\multirow{12}{*}{Gemma-4-31B}
& \multirow{4}{*}{Zero-shot}
& Overlap  & 75.11 & 75.87 & \textbf{73.73}
           & 76.39 & 78.65 & \textbf{75.78} \\
& & Conflict & 61.78 & 68.68 & \textbf{62.40}
             & 61.06 & 65.77 & \textbf{60.76} \\
& & Unique-A & 81.73 & 80.49 & \textbf{80.37}
             & 81.32 & 80.60 & \textbf{80.22} \\
& & Unique-B & 83.09 & 81.70 & \textbf{81.47}
             & 82.81 & 81.93 & \textbf{81.46} \\

\cmidrule(lr){2-9}

& \multirow{4}{*}{Few-shot}
& Overlap  & 73.90 & 76.95 & \underline{73.60}
           & 74.84 & 79.54 & \underline{75.32} \\
& & Conflict & 58.72 & 63.23 & \underline{58.11}
             & 58.33 & 60.57 & \underline{56.85} \\
& & Unique-A & 85.47 & 73.99 & \underline{77.77}
             & 85.00 & 74.07 & \underline{77.66} \\
& & Unique-B & 86.93 & 74.57 & \underline{78.54}
             & 86.60 & 74.73 & \underline{78.51} \\

\cmidrule(lr){2-9}

& \multirow{4}{*}{CoT}
& Overlap  & 70.59 & 78.75 & 72.48
           & 71.46 & 81.17 & 74.02 \\
& & Conflict & 55.46 & 62.07 & 56.13
             & 57.03 & 61.82 & 56.67 \\
& & Unique-A & 93.50 & 60.17 & 71.30
             & 93.03 & 60.26 & 71.20 \\
& & Unique-B & 92.27 & 65.18 & 74.55
             & 91.87 & 65.34 & 74.52 \\

\midrule

\multirow{16}{*}{Qwen3-8B}
& \multirow{4}{*}{Zero-shot}
& Overlap  & 50.62 & 61.95 & \underline{53.06}
           & 50.55 & 62.71 & \underline{53.34} \\
& & Conflict & 17.42 & 46.43 & 23.58
             & 17.84 & 44.93 & 23.72 \\
& & Unique-A & 66.62 & 68.88 & \underline{65.92}
             & 66.22 & 68.87 & \underline{65.68} \\
& & Unique-B & 74.01 & 68.34 & \underline{69.58}
             & 73.37 & 68.14 & \underline{69.17} \\

\cmidrule(lr){2-9}

& \multirow{4}{*}{Few-shot}
& Overlap  & 47.97 & 54.74 & 48.51
           & 48.11 & 55.50 & 48.86 \\
& & Conflict & 16.56 & 39.77 & 21.83
             & 16.88 & 38.18 & 21.90 \\
& & Unique-A & 68.97 & 55.01 & 58.16
             & 68.40 & 54.88 & 57.84 \\
& & Unique-B & 71.78 & 48.20 & 54.32
             & 70.31 & 47.92 & 53.78 \\

\cmidrule(lr){2-9}

& \multirow{4}{*}{CoT}
& Overlap  & 46.48 & 57.46 & 48.90
           & 46.26 & 58.44 & 48.92 \\
& & Conflict & 30.05 & 33.68 & \underline{29.45}
             & 32.38 & 33.35 & \underline{30.50} \\
& & Unique-A & 43.75 & 21.82 & 25.91
             & 43.75 & 21.82 & 25.91 \\
& & Unique-B & 57.92 & 37.88 & 41.67
             & 57.25 & 37.75 & 41.38 \\

\cmidrule(lr){2-9}

& \multirow{4}{*}{SFT}
& Overlap  & 70.26 & 73.14 & \textbf{70.30}
           & 70.39 & 74.54 & \textbf{71.11} \\
& & Conflict & 68.58 & 63.18 & \textbf{62.67}
             & 68.67 & 59.77 & \textbf{60.95} \\
& & Unique-A & 89.86 & 73.80 & \textbf{79.93}
             & 89.69 & 74.16 & \textbf{80.05} \\
& & Unique-B & 91.60 & 79.21 & \textbf{83.99}
             & 91.16 & 79.34 & \textbf{83.88} \\

\bottomrule
\end{tabular}

\label{tab:human_gold_comparison}
\end{table}

\subsection{Performance on Human Annotated Subset}

To examine whether our findings are sensitive to the automatically curated gold annotations, we reevaluate the methods on the human-annotated subset of 100 narrative pairs using two representative models, Gemma-4-31B and Qwen3-8B. For each model and method, we score the same predictions against both the automatically curated and human-extracted gold sets. As shown in Table~\ref{tab:human_gold_comparison}, the overall performance trends remain largely unchanged across the two evaluation settings. The largest variations occur for Overlap and Conflict, whereas the Unique tasks show only minor changes. Overall, evaluating against human-extracted annotations yields the same conclusions as the automatically curated benchmark. This suggests that the observed model and method comparisons are not driven by artifacts of the automatic curation process.


\begin{table}[!htb]
\centering
\scriptsize
\setlength{\tabcolsep}{4pt}

\caption{
Cross-domain generalization performance when the target domain is excluded from SFT training with Qwen-3-8B. Values in parentheses denote the absolute change relative to the corresponding in-domain SFT score. 
}
\label{tab:cross_domain_generalization}

\begin{tabular}{lcccc}
\toprule
\textbf{Test Domain}
& \textbf{Overlap}
& \textbf{Conflict}
& \textbf{Unique-A}
& \textbf{Unique-B} \\
\midrule

Peer
& 52.97 \bigdrop{-6.67}
& 46.29 \bigdrop{-4.79}
& 70.24 \bigdrop{-5.23}
& 70.22 \bigdrop{-5.40} \\

Factual
& 57.37 \drop{-1.59}
& 47.89 \drop{-1.86}
& 83.02 \drop{-0.23}
& 85.35 \drop{-0.71} \\

Political
& 59.99 \drop{-1.79}
& 50.41 \drop{-1.21}
& 80.21 \drop{-1.09}
& 86.07 \drop{-0.50} \\

\bottomrule
\end{tabular}

\end{table}

\begin{table}[!htb]
\centering
\small
\setlength{\tabcolsep}{5pt}

\caption{
OUC extraction under separate and joint instruction settings. Separate uses one instruction for each extraction task, whereas Joint requests all four tasks in a single instruction. Joint + SFT fine-tunes Qwen-3-8B to predict all four tasks together using a single adapter.
}
\label{tab:joint_extraction}

\begin{tabular}{llcccc}
\toprule
\textbf{Model} & \textbf{Setup}
& \textbf{Overlap}
& \textbf{Conflict}
& \textbf{Unique-A}
& \textbf{Unique-B} \\
\midrule

\multirow{3}{*}{Qwen-3-8B}
& Separate    & 40.71 & 22.72 & 63.66 & 64.48 \\
& Joint       & 33.70 & 15.88 & 41.25 & 49.70 \\
& Joint + SFT & 56.36 & 44.94 & \textbf{80.51} & \textbf{81.89} \\

\midrule

\multirow{2}{*}{Gemma-4-31B}
& Separate & \textbf{61.13} & 48.58 & 75.55 & 76.75 \\
& Joint    & 60.26 & \textbf{48.72} & 74.64 & 75.29 \\

\bottomrule
\end{tabular}

\end{table}

\begin{table*}[htb]
\centering
\tiny
\setlength{\tabcolsep}{5pt}

\caption{OUC extraction performance across prompting strategies. The highest and lowest F1-scores within each strategy are highlighted in blue and orange, respectively. The best overall score for each OUC task is shown in bold.}
\label{tab:overall_full}
\resizebox{\linewidth}{!}{%
\begin{tabular}{l|ccc|ccc|ccc|ccc}
\toprule

\multirow{2}{*}{\textbf{LLM}}
& \multicolumn{3}{c}{\textbf{Overlap}}
& \multicolumn{3}{c}{\textbf{Conflict}}
& \multicolumn{3}{c}{\textbf{Unique-A}}
& \multicolumn{3}{c}{\textbf{Unique-B}} \\

\cmidrule(lr){2-4}
\cmidrule(lr){5-7}
\cmidrule(lr){8-10}
\cmidrule(lr){11-13}

& \textbf{P} & \textbf{R} & \textbf{F1}
& \textbf{P} & \textbf{R} & \textbf{F1}
& \textbf{P} & \textbf{R} & \textbf{F1}
& \textbf{P} & \textbf{R} & \textbf{F1} \\


\midrule
\addlinespace[1pt]
\rowcolor{gray!20}
\multicolumn{13}{c}{\textbf{\textit{Zero-shot}}}\\
\addlinespace[-0.5pt]
\midrule

LLaMA-3.2-3B 
& 24.23 & 17.45 & 18.34
& 6.24 & 13.35 & 7.54
& 44.39 & 38.37 & 36.29
& 47.14 & 24.80 & 28.26 \\

Phi-4-4B
& 18.12  & 8.65 & 11.06 
& 5.96 &  6.96 & 5.90 
& 47.53 &  40.38 & 36.26
& 49.95  & 39.24 & 38.17  \\

Phi-4-14B
& 53.34 & 41.61 & 44.61
& 30.46 & 35.83 & 30.33
& 61.64 & 86.31 & \bestthree{69.72}
& 65.06 & 84.62 & \bestthree{71.64} \\

OLMo-3-7B
& 20.79 & 14.25 & 15.46
& 6.76 & 12.74 & 7.42
& 43.16 & 6.00 & \worstone{8.20}
& 39.39 & 5.71 & \worsttwo{8.56} \\

OLMo-3.1-32B
& 32.23 & 40.05 & 33.29
& 11.82 & 35.73 & 16.10
& 55.22 & 48.70 & 47.96
& 60.27 & 54.46 & 53.84 \\

Qwen-3-0.6B
& 8.48 & 2.03 & \worstone{3.04}
& 1.52 & 1.27 & \worstone{1.23}
& 43.72 & 12.62 & \worsttwo{14.20}
& 33.91 & 2.77 & \worstone{4.75} \\

Qwen-3-4B
& 35.60 & 43.12 & 36.71
& 13.32 & 38.06 & 18.17
& 56.03 & 44.87 & 45.95
& 61.90 & 52.09 & 52.97 \\

Qwen-3-8B
& 37.87 & 49.90 & 40.71
& 13.35 & 40.21 & 18.53
& 62.71 & 70.44 & 63.66
& 67.74 & 66.57 & 64.48 \\

Qwen-3-32B
& 40.45 & 61.58 & 46.77
& 25.25 & 51.97 & 31.57
& 66.83 & 64.73 & 62.39
& 72.69 & 64.87 & 65.79 \\

Qwen-3.6-35B
& 49.92 &  63.96 & \bestthree{53.42} 
& 31.64 &  56.94 & \bestthree{37.05} 
& 62.91 &  90.90 & 72.09 
& 67.57 &  88.02 & 74.54 \\

Gemma-4-2B
& 45.26 & 39.61 & 39.72
& 19.53 & 28.83 & 20.90
& 58.08 & 61.56 & 54.65
& 53.38 & 8.83 & 11.59 \\

Gemma-4-4B
& 42.83 & 57.64 & 46.41
& 23.45 & 44.83 & 28.22
& 60.41 & 86.73 & 68.43
& 61.17 & 74.83 & 61.66 \\


Gemma-4-31B
& 63.40 & 62.04 & \bestone{61.13}
& 48.90 & 54.53 & \globalbest{48.58}
& 69.08 & 87.49 & \globalbest{75.55}
& 72.68 & 84.47 & \globalbest{76.75} \\

Nemotron-3-30B
& 19.63 & 11.42 & \worsttwo{12.73}
& 6.22 & 5.68 & \worsttwo{5.11}
& 52.09 & 48.89 & 45.46
& 49.87 & 45.25 & 41.24 \\


\midrule
\addlinespace[1.5pt]
\rowcolor{gray!20}
\multicolumn{13}{c}{\textbf{\textit{Few-shot}}}\\
\addlinespace[-0.5pt]
\midrule

LLaMA-3.2-3B 
& 20.63 & 13.49 & 15.12
& 5.76 & 8.76 & 6.19
& 45.32 & 33.35 & 33.64
& 44.12 & 26.01 & 28.08 \\

Phi-4-4B
& 17.85 & 12.60 & 13.95 
& 3.55  & 4.91 & 3.68 
& 43.51 & 42.55 & 38.54
& 19.28 & 15.41 & 14.94  \\

Phi-4-14B
& 49.86 & 39.93 & 42.24
& 28.75 & 33.17 & 28.35
& 63.22 & 71.31 & 64.28
& 65.11 & 66.05 & 62.63 \\

OLMo-3-7B
& 17.22 & 15.13 & 14.74
& 5.38 & 11.67 & 6.27
& 12.60 & 7.67 & \worstone{6.93}
& 19.65 & 2.03 & \worstone{3.25} \\

OLMo-3.1-32B
& 29.61 & 35.19 & 30.07
& 4.83 & 11.20 & 5.88
& 54.74 & 40.31 & 42.69
& 58.54 & 41.73 & 45.31 \\

Qwen-3-0.6B
& 15.69 & 2.47 & \worstone{4.09}
& 1.70 & 1.08 & \worstone{1.22}
& 41.46 & 9.42 & \worsttwo{11.59}
& 37.04 & 3.51 & \worsttwo{5.66} \\

Qwen-3-4B
& 33.04 & 37.38 & 33.02
& 12.89 & 32.35 & 16.97
& 52.99 & 41.59 & 42.61
& 60.18 & 46.49 & 48.54 \\

Qwen-3-8B
& 41.13 & 41.21 & 38.78
& 14.24 & 32.22 & 18.09
& 55.77 & 53.55 & 50.91
& 61.15 & 42.62 & 46.55 \\

Qwen-3-32B
& 44.88 & 54.12 & 46.99
& 28.17 & 44.13 & 31.66
& 61.89 & 48.23 & 50.39
& 68.05 & 47.64 & 52.28 \\

Qwen-3.6-35B
& 49.66 & 64.08 & \bestthree{53.28}
& 36.68 & 50.64 & \bestthree{39.01} 
& 65.47 & 79.68 & \bestthree{69.22}
& 69.70 & 75.75 & \bestthree{70.10}  \\

Gemma-4-2B
& 43.60 & 34.53 & 36.38
& 17.55 & 24.94 & 18.22
& 17.35 & 16.43 & 14.78
& 47.84 & 17.60 & 19.26 \\

Gemma-4-4B
& 41.27 & 54.64 & 44.48
& 25.56 & 40.92 & 28.58
& 60.35 & 78.87 & 65.19
& 62.96 & 67.88 & 59.54 \\


Gemma-4-31B
& 62.27 & 63.58 & \globalbest{61.29}
& 49.92 & 52.69 & \bestone{48.27}
& 74.15 & 76.88 & \bestone{73.39}
& 77.64 & 73.73 & \bestone{73.64} \\

Nemotron-3-30B
& 18.11 & 11.21 & \worsttwo{12.43}
& 6.58 & 5.43 & \worsttwo{5.19}
& 46.89 & 43.01 & 39.79
& 46.15 & 37.40 & 35.49 \\


\midrule
\addlinespace[1.5pt]
\rowcolor{gray!20}
\multicolumn{13}{c}{\textbf{\textit{Chain-of-Thought}}}\\
\addlinespace[-0.5pt]
\midrule

LLaMA-3.2-3B 
& 22.11 & 17.25 & 17.69
& 8.45 & 10.48 & \worsttwo{7.86}
& 43.60 & 29.20 & 27.20
& 46.60 & 26.70 & 27.70 \\

Phi-4-4B
& 24.19 & 16.18 & 17.52  
& 7.26 & 6.96 & 6.22 
& 40.09 & 36.79 & 32.99
& 35.27 & 30.58 & 29.02  \\

Phi-4-14B
& 58.51 & 46.04 & 49.37
& 35.06 & 30.38 & 29.69
& 68.83 & 59.50 & 59.73
& 70.57 & 71.21 & \bestthree{67.79} \\

OLMo-3-7B
& 24.17 & 10.47 & \worsttwo{13.45}
& 14.22 & 8.56 & 9.76
& 38.81 & 8.49 & \worsttwo{11.90}
& 31.32 & 5.75 & \worsttwo{8.27} \\

OLMo-3.1-32B
& 52.81 & 46.60 & 47.39 
& 42.23 & 34.75 & 35.52
& 71.09 & 47.92 & 53.40 
& 70.18 & 52.14 & 55.57 \\

Qwen-3-0.6B
& 9.67 & 3.98 & \worstone{4.80}
& 1.06 & 1.11 & \worstone{0.08}
& 41.30 & 2.93 & \worstone{5.21}
& 32.20 & 2.30 & \worstone{4.00} \\

Qwen-3-4B
& 37.19 & 39.38 & 35.89
& 21.98 & 23.99 & 20.66
& 49.45 & 39.57 & 39.29
& 52.83 & 46.79 & 45.29 \\

Qwen-3-8B
& 38.04 & 40.11 & 37.14
& 26.22 & 24.47 & 22.72
& 30.64 & 22.30 & 22.66
& 37.96 & 29.16 & 30.01 \\

Qwen-3-32B
& 50.19 & 56.57 & 51.32
& 45.09 & 49.74 & 44.11
& 67.89 & 41.92 & 48.34
& 68.91 & 53.66 & 57.14 \\

Qwen-3.6-35B
& 61.60 & 60.88 & \bestthree{59.44}
& 46.65 & 51.81 & \bestthree{45.92} 
& 78.26 & 57.71 & 62.56 
& 58.08 & 46.47 & 49.13 \\

Gemma-4-2B
& 54.84 & 35.57 & 39.72
& 37.78 & 26.84 & 28.84
& 71.47 & 50.07 & 53.02
& 73.04 & 21.77 & 26.22 \\

Gemma-4-4B
& 61.24 & 51.65 & 54.05
& 46.37 & 39.15 & 39.47
& 69.97 & 70.02 & \besttwo{66.36}
& 71.38 & 72.46 & \besttwo{68.74} \\


Gemma-4-31B
& 60.28 & 63.49 & \bestone{60.19}
& 46.24 & 52.43 & \bestone{46.12}
& 76.79 & 56.80 & \bestthree{62.76}
& 77.94 & 62.31 & 66.98 \\

Nemotron-3-30B
& 41.60 & 39.20 & 36.50
& 20.21 & 13.46 & 14.71
& 43.96 & 26.65 & 28.59
& 47.16 & 32.14 & 33.61 \\

\bottomrule
\end{tabular}
}

\end{table*}

\subsection{Cross-Domain Generalization}
\label{app:cross-domain}
An important question is whether learned behavior via SFT generalizes beyond the domains seen during training. To evaluate this, we perform leave-one-domain-out training with Qwen-3-8B, where each target domain is excluded during SFT and used only for testing. Table~\ref{tab:cross_domain_generalization} compares this cross-domain performance with the corresponding in-domain SFT results. The results show that excluding the Factual or Political domain from training leads to only small performance drops across all four extraction tasks. In contrast, the Peer domain shows a larger drop in F1 score across all four tasks, ranging from 4.79 to 6.67. This suggests that the learned extraction behavior transfers well across the two news domains, while peer reviews appear to introduce more domain-specific variation that is harder to capture without in-domain examples.


\begin{table*}[!htbp]
\centering
\tiny
\setlength{\tabcolsep}{2.5pt}
\renewcommand{\arraystretch}{1.05}

\caption{Macro precision, recall, and F1 for different post-training strategies. Values in parentheses indicate the absolute change in F1 relative to the best prompting techniques for the corresponding model and task. Blue and red indicate improvements and degradations, respectively.}
\label{tab:overall-posttraining}

\resizebox{\linewidth}{!}{%
\begin{tabular}{lccc|ccc|ccc|ccc}
\toprule

\multirow{2}{*}{\textbf{LLM}}
& \multicolumn{3}{c}{\textbf{Overlap}}
& \multicolumn{3}{c}{\textbf{Conflict}}
& \multicolumn{3}{c}{\textbf{Unique-A}}
& \multicolumn{3}{c}{\textbf{Unique-B}} \\

\cmidrule(lr){2-4}
\cmidrule(lr){5-7}
\cmidrule(lr){8-10}
\cmidrule(lr){11-13}

& \textbf{P} & \textbf{R} & \textbf{F1}
& \textbf{P} & \textbf{R} & \textbf{F1}
& \textbf{P} & \textbf{R} & \textbf{F1}
& \textbf{P} & \textbf{R} & \textbf{F1}\\

\midrule
\addlinespace[1.5pt]
\rowcolor{gray!20}
\multicolumn{13}{c}{\textbf{\textit{Prompting Baseline}}} \\
\addlinespace[-0.5pt]
\midrule

Llama-3.2-3B
& 24.23 & 17.45 & 18.34
& 8.45 & 10.48 & 7.86
& 44.39 & 38.37 & 36.29
& 47.14 & 24.80 & 28.26 \\

Qwen-3-0.6B
& 9.67 & 3.98 & 4.80
& 1.52 & 1.27 & 1.23 
& 43.72 & 12.62 & 14.20
& 37.04  & 3.51 & 5.66 \\

Qwen-3-4B
& 35.60 & 43.12 & 36.71
& 21.98 & 23.99 & 20.66 
& 56.03 & 44.87 & 45.95
& 61.90 & 52.09 & 52.97 \\

Qwen-3-8B
& 37.87 & 49.90 & 40.71
& 26.22 & 24.47 & 22.72
& 62.71 & 70.44 & 63.66
& 67.74 & 66.57 & 64.48 \\
\midrule
\addlinespace[1.5pt]
\rowcolor{gray!20}
\multicolumn{13}{c}{\textbf{\textit{SFT}}} \\
\addlinespace[-0.5pt]
\midrule

LLaMA-3.2-3B
& 59.74 & 54.64 & 55.51\imp{37.17}
& 51.97 & 39.06 & 42.19\imp{\textbf{34.33}}
& 74.43 & 83.50 & 77.15\imp{40.86}
& 77.18 & 83.69 & 78.92\imp{50.66} \\

Qwen-3-0.6B
& 52.37 & 50.98 & 50.20\imp{\textbf{45.40}}
& 40.83 & 32.56 & 34.04\imp{32.81}
& 71.51 & 84.21 & 75.73\imp{\textbf{61.53}}
& 75.08 & 85.49 & 78.49\imp{\textbf{72.83}} \\

Qwen-3-4B
& 59.11 & 59.44 & 57.68\imp{20.97}
& 57.40 & 46.00 & 48.37\imp{27.71}
& 74.26 & 86.27 & 78.36\imp{32.41}
& 77.97 & 87.05 & 80.87\imp{27.90} \\

Qwen-3-8B
& 62.65 & 60.49 & \textbf{59.98}\imp{19.27}
& 60.79 &  47.81 & \textbf{50.82}\imp{28.10}
& 75.94 & 86.24 & \textbf{79.31}\imp{15.65}
& 79.34 & 86.70 & \textbf{81.49}\imp{17.01} \\

\midrule
\addlinespace[1.5pt]
\rowcolor{gray!20}
\multicolumn{13}{c}{\textbf{\textit{DPO}}} \\
\addlinespace[-0.5pt]
\midrule

LLaMA-3.2-3B
& 28.54 & 28.73 & 24.00\imp{5.66}
& 10.99 & 5.74 & 6.93\dec{0.93}
& 50.67 & 60.73 & 47.88\imp{11.59}
& 54.02 & 51.85 & 45.59\imp{17.33} \\

Qwen-3-0.6B
& 21.55 & 19.47 & 18.72\imp{13.92}
& 3.19 & 3.02 & 2.88\imp{1.65}
& 48.43 & 5.48 & 7.97\dec{6.23}
& 36.80 & 10.49 & 10.53\imp{4.87} \\

Qwen-3-4B
& 39.12 & 52.46 & 42.81\imp{6.10}
& 11.99 & 42.88 & 15.15\dec{5.51}
& 60.57 & 70.51 & 60.80\imp{14.85}
& 63.99 & 73.94 & 64.18\imp{11.21} \\

Qwen-3-8B
& 42.95 & 55.94 & 46.34\imp{5.63}
& 15.25 & 44.12 & 18.62\dec{4.10}
& 61.00 & 78.47 & 65.84\imp{2.18}
& 65.45 & 75.30 & 67.18\imp{2.70} \\

\midrule
\addlinespace[1.5pt]
\rowcolor{gray!20}
\multicolumn{13}{c}{\textbf{\textit{GRPO}}} \\
\addlinespace[-0.5pt]
\midrule

LLaMA-3.2-3B
& 23.94 & 48.16 & 28.74\imp{10.40}
& 19.98 & 28.73 & 21.19\imp{13.33}
& 60.53 & 88.53 & 69.95\imp{33.66}
& 66.73 & 83.57 & 72.40\imp{44.14} \\

Qwen-3-0.6B
& 26.75 & 26.61 & 25.36\imp{20.56}
& 4.43 & 9.43 & 5.52\imp{4.29}
& 57.11 & 93.16 & 68.76\imp{54.56}
& 56.61 & 90.34 & 67.52\imp{61.86} \\

Qwen-3-4B
& 57.04 & 48.57 & 50.74\imp{14.03}
& 40.71 & 33.64 & 33.89\imp{13.23}
& 65.86 & 86.52 & 72.92\imp{26.97}
& 69.79 & 86.65 & 75.65\imp{22.68} \\

Qwen-3-8B
& 63.71 &	52.03 & 55.58\imp{14.87}
& 43.35 & 41.89 & 39.45\imp{16.73}
& 67.70 & 87.52	& 74.60\imp{10.94}
& 72.62 & 86.53 & 77.47\imp{12.99}\\

\bottomrule
\end{tabular}
}

\end{table*}

\subsection{Joint OUC Extraction}
\label{app:joint-ouc}
So far, we have used a separate instruction for each OUC task, allowing the model to focus on one relation at a time rather than distinguish and extract all four within a single response. This setup, however, requires four model calls for each narrative pair. To study whether the four outputs can be produced more efficiently, we also evaluate a joint instruction that requests all OUC outputs at once. Table~\ref{tab:joint_extraction} reports results for Qwen-3-8B and Gemma-4-31B. We choose Qwen-3-8B because it performs best among the models used in our post-training experiments, making it our primary candidate for joint fine-tuning. Gemma-4-31B is included as a reference because it achieves the best prompting performance overall. The comparison is therefore not intended as a controlled study of model size or family. The results show that under joint instruction, Qwen-3-8B drops across all four tasks, whereas Gemma-4-31B remains close to its performance with separate instructions. This shows that the ability to handle all four extraction tasks within a single response varies considerably across models. We therefore fine-tune Qwen-3-8B jointly with a single LoRA adapter, which substantially improves performance and recovers much of the loss observed under joint instruction. However, the performance is still below that of Gemma-4-31B on Overlap and Conflict, but surpasses it on both Unique tasks. Compared with a separate task-specific SFT, joint SFT sacrifices some performance, yet it reduces four inference calls to one and requires only a single adapter. Thus, joint fine-tuning provides a more efficient alternative when a small reduction in performance is acceptable.













\begin{table*}[t]
\centering
\tiny
\setlength{\tabcolsep}{3pt}
\renewcommand{\arraystretch}{1.15}

\caption{
Representative examples from five error categories. For Wrong Pairing, the predicted counterpart is shown together with the corresponding gold counterpart.
}
\label{tab:error_examples}

\begin{tabular}{p{2.0cm} p{1.0cm} p{5.7cm} p{5.3cm}}
\toprule
\textbf{Error Type} & \textbf{Task} & \textbf{Example} & \textbf{Explanation} \\
\midrule

\multirow{3}{2.0cm}{\textbf{Missed Information}}
& Overlap
& A: ``18 political parties are taking part in the election process.'' \newline
B: ``A total of 18 political parties contested the election.''
& The two clauses express the same fact, but the model does not extract the pair. \\

& Overlap
& A: ``The possibility of synchronously operating multiple viewports, such as state and reward, is highlighted as a key strength.'' \newline
B: ``The possibility of synchronously operating multiple viewports, such as state and reward, is highlighted as a key strength.''
& The clauses are identical, but the model misses the Overlap pair. \\

& Conflict
& A: ``By removing term limits, the constitution now effectively guarantees Xi Jinping can remain president for life.'' \newline
B: ``Party officials stress that the removal of term limits does not lock any individual in office for life.''
& The two clauses make opposing claims about the consequence of removing term limits, but the model misses the pair. \\

\midrule

\multirow{3}{2.0cm}{\textbf{Wrong Semantic Match}}
& Conflict
& A: ``Polls show a sizable portion of the Chinese public is skeptical or opposed to the constitutional change.'' \newline
B: ``The party's official People's Daily reprinted a long article by Xinhua news agency saying most people supported the constitutional amendments, quoting a variety of people proffering support.''
& A sizable opposing group can coexist with majority support, so the statements are not contradictory. \\

& Conflict
& A: ``The head of the Catholic Church in the Philippines has harshly criticized a government campaign of alleged extrajudicial killings of drug suspects that has claimed thousands of lives, calling it a `humanitarian concern' that cannot be ignored.'' \newline
B: ``Most religious leaders have stayed out of the debate, offering no public criticism of the anti-drug drive.''
& Criticism from one religious leader does not contradict a claim about the behavior of most religious leaders. \\

& Conflict
& A: ``Beijing has already provided the former Soviet republic with loans worth \$10bn, but the government must find more than \$17bn in 2014 to meet gas bills and debt repayments.'' \newline
B: ``Despite talks, China has not yet granted Ukraine any new loans.''
& Previously provided loans and the absence of new loans can both be true, so the pair does not express a contradiction. \\

\midrule

\multirow{3}{2.0cm}{\textbf{Wrong Pairing}}
& Conflict
& Predicted A: ``Somali pirates who seized a Comoros-flagged oil tanker earlier this week after five years without a major hijacking in the region have released the ship and its crew without conditions, officials said late Thursday.'' \newline
B: ``Families of the eight Sri Lankan crew members held captive by Somali pirates on an oil tanker tearfully pleaded Wednesday for the men to be released unharmed, while the pirates demanded a ransom.'' \newline
Gold A: ``The pirates told authorities that the only reason they seized the ship was in protest of the illegal fishing in the area that has threatened livelihoods, not for ransom, Mohamed said.''
& The release of the ship does not contradict the earlier ransom demand. The actual Conflict concerns whether the seizure was for ransom. \\

& Conflict
& Predicted A: ``Reporting RGB-based PSNR is perfectly valid for comparing against H.264, as the authors have done.'' \newline
Predicted B: ``In the video compression literature NOBODY reports RGB reconstruction metrics.'' \newline
Gold B: ``Please note that video codecs DO NOT OPTIMIZE FOR RGB reconstruction ... so comparing against them in that color space puts them at a distinct disadvantage.''
& The predicted counterpart concerns reporting practice, whereas the gold counterpart directly challenges the validity of the RGB comparison. \\

& Overlap
& Predicted A: ``He said doctors had to pick sharp, hardened fragments of lava out of the wound, but the prognosis is good for his friend.'' \newline
B: ``It hit him on the shin and shattered everything there down on his leg,'' she said, adding that lava spatters ``can weigh as much as a refrigerator and even small pieces of spatter can kill.'' \newline
Gold A: ``The eruptions caused the first known serious injury on the island.''
& The predicted clause describes subsequent treatment, rather than the injury event that corresponds to the second clause. \\

\midrule

\multirow{3}{2.0cm}{\textbf{Span / Verbatim Error}}
& Overlap
& Model output: ``Russia wants to draw Ukraine into a Moscow-led customs union and prevent it drawing closer to the EU, a move that would signal a historic shift towards the West and away from Kiev's former Soviet masters in Moscow.''
& The model expresses relevant information, but the sentence is not an exact span from the source narrative. \\

& Conflict
& Model output A: ``The reported memory usage for ConvOcc is clearly erroneous.'' \newline
Model output B: ``The reported memory usage for ConvOcc appears consistent with its known constant footprint.''
& The model generates reformulated statements instead of extracting the original review sentences verbatim. \\

& Overlap
& Model output A: ``The approach of importing reinforcement-learning skill discovery into an educational context is a fresh perspective that could inspire further cross-disciplinary work.'' \newline
Model output B: ``Introducing RL-derived skill primitives as building blocks for human tutoring indeed represents an innovative cross-domain contribution.''
& The model paraphrases the relevant information rather than returning exact source spans. \\

\midrule

\multirow{3}{2.0cm}{\textbf{Relation Confusion}}
& Overlap
& A: ``The baseline suite includes recent VAE-based detectors such as EGBAD and AnoGAN, demonstrating that the proposed method outperforms state-of-the-art unsupervised approaches.'' \newline
B: ``The experimental comparison omits recent VAE-based baselines such as EGBAD and AnoGAN, making it difficult to gauge relative performance.''
& The model predicts Overlap, although one clause says the baselines are included and the other says they are omitted. \\

& Overlap
& A: ``This year Hungary has recorded around 18,000 illegal border crossings.'' \newline
B: ``Police reported only about 5,000 illegal border crossings this year, far fewer than the 18,000 recorded last year.''
& The model predicts Overlap even though the clauses give incompatible values for the same year. \\

& Conflict
& A: ``The 1916 Giants went 27 games without a loss but had a tie mixed in.'' \newline
B: ``The 1916 New York Giants won 26 straight games without a loss, with a tie included in the run.''
& The model predicts Conflict, although the statements are compatible: 26 wins and one tie constitute a 27-game unbeaten run. \\

\bottomrule
\end{tabular}

\end{table*}

\subsection{Data Generation Prompts}
\label{app:prompt-data-generation}
\begin{promptbox}{Overlap Candidate Extraction}
You are an extractor that compares two multi-perspective narratives about the same topic and extracts sentence-level OVERLAP pairs according to the provided task definitions and output format.

\textbf{DEFINITIONS}

\textbf{Aspect:} A short phrase naming the specific claim or opinion both sentences address. Must be specific enough that both sentences answer the same question---not a broad topic like ``the paper'' or ``the algorithm'' alone.

\textbf{Overlap pair:} Same aspect; both sentences express compatible claims or compatible opinions---both can be true simultaneously. Includes paraphrases and compatible evaluations on the same point.

\textbf{READING INSTRUCTIONS}

Before extracting any pairs, read both Narratives completely from start to finish. For each sentence in Narrative 1, ask whether any sentence in Narrative 2 addresses the same specific aspect in a compatible way. Then do the same starting from Narrative 2. Do not rely on surface similarity alone. Implicit overlaps require understanding what each sentence is actually claiming, not just matching keywords.

\textbf{RULES}

\begin{enumerate}[leftmargin=*]
    \item Same specific aspect only. Both sentences must address the same
    specific claim or opinion---not just the same topic, event, person,
    or paper.

    \item Compatibility test: can both sentences be true simultaneously?
    If one affirms what the other denies, it is NOT overlap — skip it.

    \item One sentence from each Narrative per pair. Never pair two sentences
    from the same Narrative.

    \item Extract only complete sentence-level units. Do not return partial spans or sentence fragments.

    \item Verbatim only. Copy sentences exactly as they appear in the input.

    \item Identical sentences are valid explicit overlap pairs.

    \item A sentence may appear in multiple pairs only when the aspects
    genuinely differ.

    \item Prefer precision over recall. When in doubt, skip. An empty array
    is better than a wrong pair.
\end{enumerate}

\textbf{OUTPUT:} Single JSON object only, no commentary:

\begin{verbatim}
{
  "overlap_pairs": [
    {
      "aspect": "...",
      "sentence_doc1": "...",
      "sentence_doc2": "..."
    }
  ]
}
\end{verbatim}

Now read both Narratives carefully and completely before extracting overlap pairs.

Narrative 1:

\texttt{\_\_Narrative1\_\_}

Narrative 2:

\texttt{\_\_Narrative2\_\_}

Return only the JSON object with \texttt{overlap\_pairs}.
\end{promptbox}

\begin{promptbox}{Conflict Candidate Extraction}
You are an extractor that compares two multi-perspective narratives about the same topic and extracts sentence-level CONFLICT pairs according to the provided task definitions and output format.

\textbf{DEFINITIONS}

\textbf{Aspect:} A short phrase naming the specific claim or opinion both sentences address. Must be specific enough that both sentences answer the same question---not a broad topic like ``the paper'' or ``the algorithm'' alone.

\textbf{Conflict pair:} Same aspect; both sentences express incompatible claims or
opposing opinions---both cannot be true simultaneously. A reliable test: could a single person assert both sentences about the same thing without contradicting themselves? If yes, it is not a conflict. If no, it is.

\textbf{Common conflict patterns:}
\begin{itemize}[leftmargin=*,nosep]
    \item One sentence says X is true, the other says X is false.
    \item One sentence gives metric value A, the other gives incompatible value B.
    \item One sentence praises a specific aspect, the other criticizes that
    same specific aspect.
    \item One sentence characterizes an action as justified, the other as harmful.
\end{itemize}

\textbf{Common non-conflict patterns to avoid:}
\begin{itemize}[leftmargin=*,nosep]
    \item Two sentences both criticizing different aspects of the same subject.
    \item Two sentences raising concerns about different specific points.
    \item One sentence about topic X, another about related but distinct topic Y.
\end{itemize}

\textbf{READING INSTRUCTIONS}

Before extracting any pairs, read both Narratives completely from start to finish. For each sentence in Narrative 1, ask whether any sentence in Narrative 2 addresses the same specific aspect in an incompatible way. Then do the same starting from Narrative 2. Do not rely on surface similarity alone. Implicit conflicts require understanding what each sentence is actually claiming.

\textbf{RULES}

\begin{enumerate}[leftmargin=*]
    \item Same specific aspect only. Both sentences must address the same
    specific claim or opinion---not just the same topic, event, person,
    or paper.

    \item Self-contradiction test: could a single person assert both sentences
    about the same thing without contradicting themselves? If yes, skip.

    \item One sentence from each Narrative per pair. Never pair two sentences
    from the same Narrative.

    \item Extract only complete sentence-level units. Do not return partial spans or sentence fragments.

    \item Verbatim only. Copy sentences exactly as they appear in the input.

    \item A sentence may appear in multiple pairs only when the aspects
    genuinely differ.

    \item Prefer precision over recall. When in doubt, skip. An empty array
    is better than a wrong pair.
\end{enumerate}

\textbf{OUTPUT:} Single JSON object only, no commentary:

\begin{verbatim}
{
  "conflict_pairs": [
    {
      "aspect": "...",
      "sentence_doc1": "...",
      "sentence_doc2": "..."
    }
  ]
}
\end{verbatim}

Read both Narratives carefully and completely before extracting conflict pairs.

Narrative 1:

\texttt{\_\_Narrative1\_\_}

Narrative 2:

\texttt{\_\_Narrative2\_\_}

Return only the JSON object with \texttt{conflict\_pairs}.
\end{promptbox}

\begin{promptbox}{Overlap and Conflict Validation Prompt}
You are a strict validator of one sentence-level
\texttt{\_\_CLAIMED\_TYPE\_\_} pair extracted from two
multiple-perspective narratives. Assign a verdict to this single pair
using the pair itself and the two source narratives.

\textbf{DEFINITIONS}

\textbf{Aspect:} A short phrase naming the specific claim or opinion both sentences address. Must be specific enough that both sentences answer the same question---not a broad topic like ``the paper'' or ``the algorithm'' alone.

\textbf{Overlap:} Same aspect; both sentences express compatible claims or compatible opinions---both can be true simultaneously. Includes paraphrases and compatible evaluations on the same point.

\textbf{Conflict:} Same aspect; both sentences express incompatible claims or incompatible opinions---they cannot both be true simultaneously. A reliable test: could a single person assert both sentences about the same thing without contradicting themselves? If yes, not a conflict.

\textbf{Invalid:} The pair has no valid relationship---the sentences address different aspects, one is a fragment, both come from the same narrative, or there is no genuine overlap or conflict.

\textbf{RULES}

\begin{enumerate}[leftmargin=*]
    \item Read both source narratives before validating the pair. Use the
    narratives only to understand the context, referents, and intended
    meaning of the two sentences.

    \item Same specific aspect only. Both sentences must answer the same
    specific question. If they share a topic but address different
    questions, verdict is \texttt{"invalid"}.

    \item Complete sentences only. A fragment or clause that depends on
    surrounding context for its meaning gets verdict \texttt{"invalid"}.

    \item Identical \texttt{sentence\_1} and \texttt{sentence\_2} are always valid overlaps — do not mark them invalid.

    \item For claimed overlap---compatibility test: can both sentences be
    true simultaneously? If yes, verdict is \texttt{"overlap"}. If they
    contradict each other on the same specific claim, verdict is
    \texttt{"conflict"}. Only flag as conflict if \texttt{sentence\_2}
    directly negates or contradicts the specific claim made in
    \texttt{sentence\_1}---additional commentary, framing, or skepticism
    about a secondary point does not make a pair a conflict.

    \item For claimed conflict---self-contradiction test: could a single person assert both sentences without contradicting themselves?
    If no, verdict is \texttt{"conflict"}. If yes, verdict is
    \texttt{"overlap"}.

    \item Reclassification is preferred over discarding. Only use
    \texttt{"invalid"} when the pair has no valid relationship or the
    aspect is not specific enough at all.
\end{enumerate}

\textbf{OUTPUT:} Single JSON object only, no commentary:

\begin{verbatim}
{"verdict": "overlap"}
\end{verbatim}

\texttt{verdict} must be exactly one of:
\texttt{"overlap"}, \texttt{"conflict"}, \texttt{"invalid"}.
\end{promptbox}

\begin{promptbox}{User Prompt}
Read both source narratives and validate this pair. It was extracted as
\texttt{\_\_CLAIMED\_TYPE\_\_}.

\textbf{Source Narrative 1:}

\texttt{\_\_narrative1\_\_}

\textbf{Source Narrative 2:}

\texttt{\_\_narrative2\_\_}

\textbf{Pair to validate:}

\texttt{\_\_PAIR\_JSON\_\_}

Return \texttt{\{"verdict": "..."\}} only.
\end{promptbox}

\begin{promptbox}{Unique Clause Extraction (Narrative 1)}
You are an extractor that compares two multi-perspective narratives about the same topic and extracts UNIQUE sentences from Narrative 1 relative to Narrative 2. You are given both Narratives, along with the already-extracted overlap and conflict pairs.

\textbf{DEFINITIONS}

\textbf{Overlap pair:} Two sentences (one from each Narrative) that address the same specific aspect with compatible meaning---both can be true simultaneously.

\textbf{Conflict pair:} Two sentences (one from each Narrative) that address the same specific aspect with incompatible meaning---both cannot be true simultaneously.

\textbf{Unique sentence (Narrative 1):} A complete sentence from Narrative 1 for which Narrative 2 is completely silent on the same specific fact, event, situation,
or claim---whether with compatible or incompatible meaning.

A Narrative 1 sentence is NOT unique if:
\begin{itemize}[leftmargin=*,nosep]
    \item It appears in any provided overlap pair as \texttt{sentence\_doc1}.
    \item It appears in any provided conflict pair as \texttt{sentence\_doc1}.
    \item Any sentence in Narrative 2 addresses the same specific claim, even with
    different wording (paraphrase, question, agreement, or contradiction).
\end{itemize}

\textbf{READING INSTRUCTIONS}

\begin{enumerate}[leftmargin=*]
    \item Read Narrative 1 and Narrative 2 completely from start to finish.
    \item Read all provided \texttt{overlap\_pairs} and \texttt{conflict\_pairs}.
    \item For each sentence in Narrative 1, decide whether Narrative 2 (including
    sentences already paired in overlap/conflict lists) addresses the same
    specific claim. If yes, exclude it. If Narrative 2 is silent on that
    specific claim, include it as unique.
\end{enumerate}

Do not rely on surface wording alone. Paraphrases and implicit counterparts count as non-unique even when wording differs completely.

\textbf{RULES}

\begin{enumerate}[leftmargin=*]
    \item Source side only. Every unique item must be a verbatim sentence from
    Narrative 1. \texttt{sentence\_doc2} must always be the empty string \texttt{""}.

    \item Cross-check pairs first. Exclude any Narrative 1 sentence that appears
    in \texttt{overlap\_pairs} or \texttt{conflict\_pairs} as
    \texttt{sentence\_doc1}.

    \item Cross-check Narratives second. Even if a sentence is absent from the
    pair lists, exclude it when any Narrative 2 sentence addresses the same
    specific claim (overlap or conflict).

    \item Extract only complete sentence-level units. Do not return partial spans or sentence fragments.

    \item Verbatim only. Copy sentences exactly as they appear in Narrative 1.

    \item Specific claim, not broad topic. Two sentences about the same paper,
    person, or event are not automatically counterparts unless they address
    the same specific fact or claim.
\end{enumerate}

\textbf{OUTPUT:} Single JSON object only, no commentary:

\begin{verbatim}
{
  "unique1_items": [
    {
      "sentence_doc1": "...",
      "sentence_doc2": ""
    }
  ]
}
\end{verbatim}

Now read both Narratives and the provided overlap/conflict pairs before extracting unique sentences from Narrative 1.

Narrative 1:

\texttt{\_\_Narrative1\_\_}

Narrative 2:

\texttt{\_\_Narrative2\_\_}

Overlap pairs already extracted between these Narratives:

\texttt{\_\_OVERLAP\_PAIRS\_\_}

Conflict pairs already extracted between these Narratives:

\texttt{\_\_CONFLICT\_PAIRS\_\_}

Return only the JSON object with \texttt{unique1\_items}.
\end{promptbox}

\begin{promptbox}{Unique Clause Extraction (Narrative 2)}
You are an extractor that compares two multi-perspective narratives about the same topic and extracts UNIQUE sentences from Narrative 2 relative to Narrative 1. You are given both Narratives, along with the already-extracted overlap and conflict pairs.

\textbf{DEFINITIONS}

\textbf{Overlap pair:} Two sentences (one from each Narrative) that address the same specific aspect with compatible meaning---both can be true simultaneously.

\textbf{Conflict pair:} Two sentences (one from each Narrative) that address the same specific aspect with incompatible meaning---both cannot be true simultaneously.

\textbf{Unique sentence (Narrative 2):} A complete sentence from Narrative 2 for which Narrative 1 is completely silent on the same specific fact, event, situation, or claim: whether with compatible or incompatible meaning.

A Narrative 2 sentence is NOT unique if:
\begin{itemize}[leftmargin=*,nosep]
    \item It appears in any provided overlap pair as \texttt{sentence\_doc2}.
    \item It appears in any provided conflict pair as \texttt{sentence\_doc2}.
    \item Any sentence in Narrative 1 addresses the same specific claim, even with
    different wording (paraphrase, question, agreement, or contradiction).
\end{itemize}

\textbf{READING INSTRUCTIONS}

\begin{enumerate}[leftmargin=*]
    \item Read Narrative 1 and Narrative 2 completely from start to finish.
    \item Read all provided \texttt{overlap\_pairs} and \texttt{conflict\_pairs}.
    \item For each sentence in Narrative 2, decide whether Narrative 1 (including
    sentences already paired in overlap/conflict lists) addresses the same
    specific claim. If yes, exclude it. If Narrative 1 is silent on that
    specific claim, include it as unique.
\end{enumerate}

Do not rely on surface wording alone. Paraphrases and implicit counterparts count as non-unique even when wording differs completely.

\textbf{RULES}

\begin{enumerate}[leftmargin=*]
    \item Source side only. Every unique item must be a verbatim sentence from
    Narrative 2. \texttt{sentence\_doc1} must always be the empty string \texttt{""}.

    \item Cross-check pairs first. Exclude any Narrative 2 sentence that appears
    in \texttt{overlap\_pairs} or \texttt{conflict\_pairs} as
    \texttt{sentence\_doc2}.

    \item Cross-check Narratives second. Even if a sentence is absent from the
    pair lists, exclude it when any Narrative 1 sentence addresses the same
    specific claim (overlap or conflict).

    \item Extract only complete sentence-level units. Do not return partial spans or sentence fragments.

    \item Verbatim only. Copy sentences exactly as they appear in Narrative 2.

    \item Specific claim, not broad topic. Two sentences about the same paper, person, or event are not automatically counterparts unless they address the same specific fact or claim.

    \item Prefer precision over recall. When in doubt, skip. An empty array is
    better than a wrong sentence.
\end{enumerate}

\textbf{OUTPUT:} Single JSON object only, no commentary:

\begin{verbatim}
{ 
    "unique2_items": [
    {
      "sentence_doc1": "",
      "sentence_doc2": "..."
    }]
}
\end{verbatim}

Now, read both Narratives and the provided overlap/conflict pairs before extracting unique sentences from Narrative 2.

Narrative 1:

\texttt{\_\_Narrative1\_\_}

Narrative 2:

\texttt{\_\_Narrative2\_\_}

Overlap pairs already extracted between these Narratives:

\texttt{\_\_OVERLAP\_PAIRS\_\_}

Conflict pairs already extracted between these Narratives:

\texttt{\_\_CONFLICT\_PAIRS\_\_}

Return only the JSON object with \texttt{unique2\_items}.
\end{promptbox}

\subsection{Annotation Guidelines}
\label{app:guidelines}

We annotate candidate outputs for the three OUC relations using two labels: \textbf{Valid} and \textbf{Invalid}. For Overlap and Conflict, the annotation unit is a pair of clauses, one from each narrative. For Unique-A and Unique-B, the annotation unit is a single clause from one narrative. Annotators read both narratives before making a decision and judge each candidate based on its meaning in context rather than lexical similarity alone.

\paragraph{Overlap.}
A candidate pair is labeled \textbf{Valid} when the two clauses refer to the same underlying information and express compatible meanings. The wording may differ, and one clause may contain slightly more detail, as long as the shared information remains the same. A pair is labeled \textbf{Invalid} when the clauses are only topically related, describe different aspects of the same event or entity, or express incompatible claims.

For example,
\begin{quote}
\small
\textit{``Thousands of people gathered downtown for the protest.''}\\
\textit{``The demonstration drew thousands of participants to the city center.''}
\end{quote}
is a valid Overlap pair because both clauses convey the same underlying
information. In contrast,
\begin{quote}
\small
\textit{``The paper does not define several technical terms.''}\\
\textit{``The organization of the paper makes it difficult to follow.''}
\end{quote}
is invalid because the two clauses discuss different aspects of the paper,
despite being broadly related.

\paragraph{Conflict.}
A candidate pair is labeled \textbf{Valid} when the two clauses refer to the same underlying information but make incompatible claims about it. The disagreement may concern a fact, outcome, interpretation, quantity, or other property, provided that both clauses address the same point. A pair is labeled \textbf{Invalid} when the clauses concern different information, describe different aspects, or can both be true without contradiction.

For example,
\begin{quote}
\small
\textit{``The policy significantly reduced emissions.''}\\
\textit{``The policy produced no measurable reduction in emissions.''}
\end{quote}
is a valid Conflict pair because both clauses address the same effect of the
policy but make incompatible claims. In contrast,
\begin{quote}
\small
\textit{``The paper is clearly written.''}\\
\textit{``The experimental section lacks sufficient detail.''}
\end{quote}
is invalid because the two statements evaluate different aspects and are not
directly contradictory.

\begin{figure}[!b]
    \centering
    \includegraphics[width=0.9\linewidth]{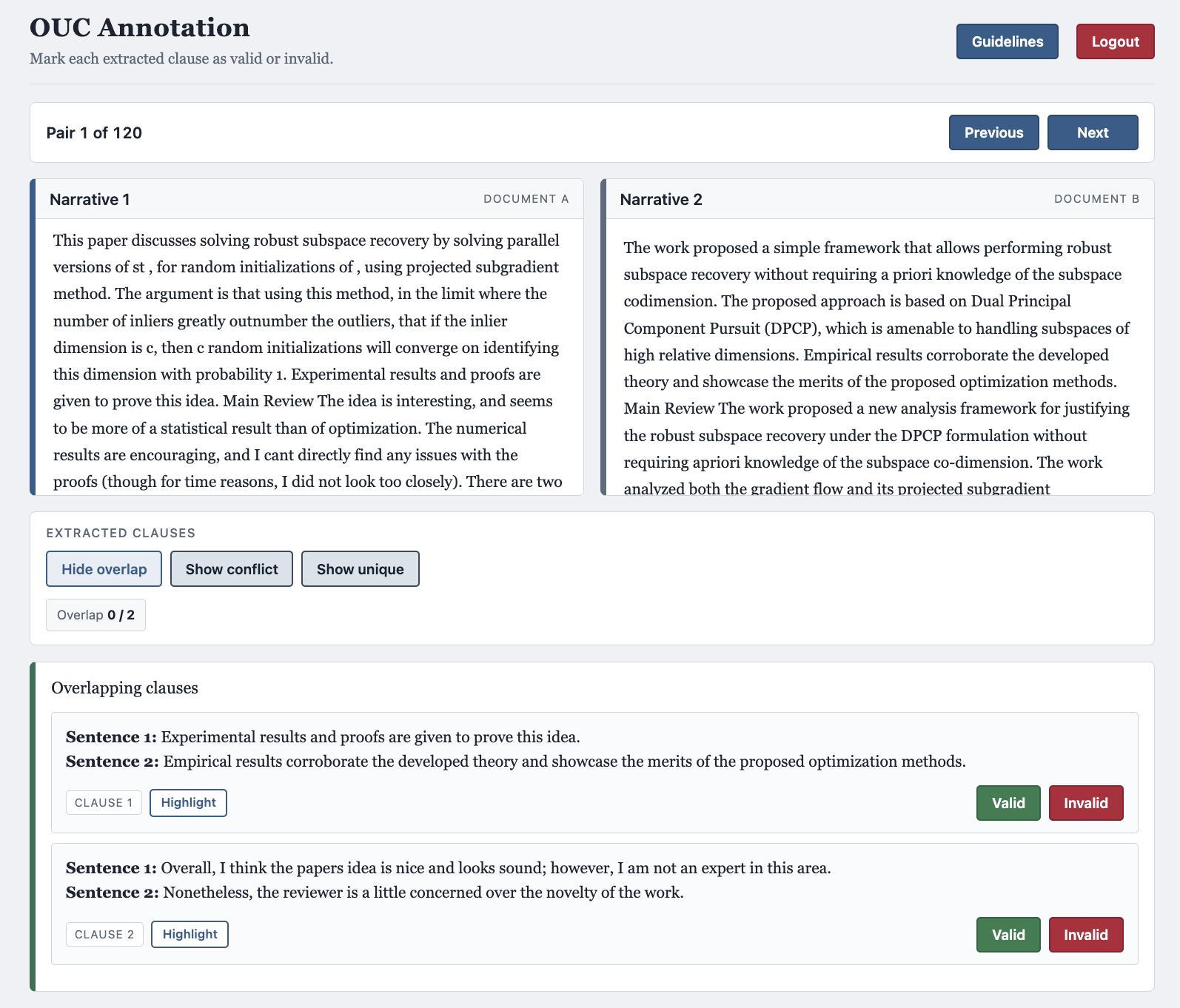}
    \caption{Annotation Interface}
    \label{fig:annotation-tool}
\end{figure}

\paragraph{Unique.}
A clause is labeled \textbf{Valid} for Unique-A or Unique-B when the information it expresses has no corresponding information anywhere in the alternative narrative. A clause is labeled \textbf{Invalid} if the alternative narrative contains a counterpart expressing the same underlying information, whether that counterpart agrees with or conflicts with the clause.

For example, if Narrative A states
\begin{quote}
\small
\textit{``The mayor announced that the bridge would reopen on Monday.''}
\end{quote}
and Narrative B contains no information about the bridge reopening, the clause is valid Unique-A. However, if Narrative B states that the bridge would reopen on Tuesday, the clause is not Unique because the same underlying information is present in both narratives, even though the two claims conflict.

\paragraph{General Decision Criteria.}
Annotators follow three main principles. First, sharing a broad topic is not sufficient for either Overlap or Conflict; the clauses must refer to the same underlying information. Second, differences in wording alone do not make a pair invalid. Third, any information that has a corresponding Overlap or Conflict counterpart in the alternative narrative is not considered Unique.

\paragraph{Annotation Interface.}
To facilitate annotation, we developed a lightweight web-based interface that presents the candidate clause or clause pair together with the two complete narratives. Annotators can label each candidate as \textbf{Valid} or \textbf{Invalid} and, when needed, highlight the corresponding clause spans
directly within the full narratives to verify that the extracted text is correctly grounded in context. Figure~\ref{fig:annotation-tool} shows a snapshot of the annotation interface.

\subsection{Evaluation Prompts}
\label{app:prompt-inference}

\subsubsection{Zero-shot Prompt}

\begin{promptbox}{Overlap}
You are given two multi-perspective narratives: Narrative 1 and Narrative 2. Extract all pairs of sentences (one from Narrative 1 and one from Narrative 2) that overlap---that is, both sentences describe the same specific fact, event, situation, or claim and express compatible meaning (both can be true at the same time). Read both narratives carefully and extract all pairs of sentences that overlap.

\textbf{Rules:}
\begin{itemize}[leftmargin=*,nosep]
    \item Do NOT include pairs with incompatible or contradictory meanings.
    \item Do NOT match sentences based only on broad topic similarity.
    \item The sentences must refer to the same specific fact, event, or claim or opinion or take same stance on the same aspect.
    \item Do NOT include sentence pairs that are not overlapping when evaluated under the same aspect.
    \item If unsure, exclude the pair.
\end{itemize}

\textbf{Input:}

Narrative 1: \texttt{\_\_Narrative1\_\_}

Narrative 2: \texttt{\_\_Narrative2\_\_}

Respond with a JSON object with a single key \texttt{"pairs"}, where each item is:

\begin{verbatim}
{
  "sentence_1": string,
  "sentence_2": string
}
\end{verbatim}

If there are no valid pairs, return:

\begin{verbatim}
{"pairs": []}
\end{verbatim}

Return ONLY valid JSON. No explanation.
\end{promptbox}

\begin{promptbox}{Conflict}
You are given two narratives: Narrative 1 and Narrative 2. Each narrative consists of multiple sentences. Extract all conflicting pairs of sentences (one from Narrative 1 and one from Narrative 2)---that is, both sentences refer to the same specific fact, event, or claim that make clearly opposing or contradictory statements that cannot both be true. In other words, both sentences take different stances on the same aspect or make clearly opposing or contradictory statements that cannot both be true on the same aspect. Read both narratives carefully and extract all pairs of sentences that conflict.

\textbf{Rules:}
\begin{itemize}[leftmargin=*,nosep]
    \item Do NOT include pairs with compatible meanings such those are similar on the same aspect but not contradictory statements.
    \item Do NOT include sentence pairs that doesn't contradict when evaluated under the same aspect.
    \item Include only valid conflicting pairs.
\end{itemize}

\textbf{Input:}

Narrative 1: \texttt{\_\_Narrative1\_\_}

Narrative 2: \texttt{\_\_Narrative2\_\_}

Respond with a JSON object with a single key \texttt{"pairs"}, where each item is:

\begin{verbatim}
{
  "sentence_1": string,
  "sentence_2": string
}
\end{verbatim}

If there are no such pairs, return:

\begin{verbatim}
{"pairs": []}
\end{verbatim}

Return ONLY valid JSON. No explanation.
\end{promptbox}

\begin{promptbox}{Unique-A}
You are given two narratives about the same topic, possibly from different perspectives. Identify all sentences from Narrative 1 that are unique relative to Narrative 2---that is, no sentence in Narrative 2 refers to the same specific fact, event, situation, or claim, whether with compatible (overlapping) or incompatible (conflicting) meaning.

\textbf{Input:}

Narrative 1: \texttt{\_\_Narrative1\_\_}

Narrative 2: \texttt{\_\_Narrative2\_\_}

Respond with a JSON object with a single key \texttt{"items"} whose value is an array of objects. EVERY object MUST include both keys:

\texttt{"sentence\_1"}: string (the unique sentence from Narrative 1), a single sentence, not multiple sentences.

\texttt{"sentence\_2"}: always the empty string \texttt{""}.

Example:

\begin{verbatim}
{"items":[{"sentence_1":"string",
"sentence_2":""}]}
\end{verbatim}

If there are no unique sentences, return:

\begin{verbatim}
{"items": []}
\end{verbatim}

Return ONLY valid JSON. No explanation.
\end{promptbox}

\begin{promptbox}{Unique-B}
You are given two narratives about the same topic that present different perspectives. Identify all sentences from Narrative 2 that are unique relative to Narrative 1---that is, no sentence in Narrative 1 refers to the same specific fact, event, situation, or claim, whether with compatible (overlapping) or incompatible (conflicting) meaning.

\textbf{Input:}

Narrative 1: \texttt{\_\_Narrative1\_\_}

Narrative 2: \texttt{\_\_Narrative2\_\_}

Respond with a JSON object with a single key \texttt{"items"} whose value is an array of objects.

EVERY object MUST include both keys:

\texttt{"sentence\_1"}: always the empty string \texttt{""}.

\texttt{"sentence\_2"}: string (the unique sentence from Narrative 2), a single sentence, not multiple sentences.

Example:

\begin{verbatim}
{"items":[{"sentence_1":"",
"sentence_2":"string"}]}
\end{verbatim}

If there are no unique sentences, return:

\begin{verbatim}
{"items": []}
\end{verbatim}

Return ONLY valid JSON. No explanation.
\end{promptbox}

\subsubsection{Few-shot Prompt}

\begin{promptbox}{Overlap}
You are given two multi-perspective narratives: Narrative 1 and Narrative 2. Identify all pairs of sentences (one from Narrative 1 and one from Narrative 2) that overlap: that is, both sentences describe the same specific fact, event, situation, or claim and express compatible meaning (both can be true at the same time).

\textbf{Rules:}
\begin{itemize}[leftmargin=*,nosep]
    \item Do NOT include pairs with incompatible or contradictory meanings.
    \item Do NOT match sentences based only on broad topic similarity.
    \item The sentences must refer to the same specific fact, event, or claim.
    \item Do NOT include weak or vague similarities.
    \item If unsure, exclude the pair.
    \item A sentence may match multiple sentences.
\end{itemize}

\textbf{Examples of Valid Overlap Pairs:}

\{\{6-shot examples\}\}

\textbf{Examples of Pairs That Look Similar but Are Not Overlap:}

\{\{4 counter-examples\}\}

Now apply the same standard to the narratives below. Before including a pair, always check: (1) do the two sentences refer to the same specific fact, event, or claim---not just the same broad topic or person, and (2) can both sentences be true at the same time? If either check fails, exclude the pair.

\textbf{Input:}

Narrative 1: \texttt{\_\_Narrative1\_\_}

Narrative 2: \texttt{\_\_Narrative2\_\_}

Respond with a JSON object with a single key \texttt{"pairs"}, where each item is:

\begin{verbatim}
{
  "sentence_1": string,
  "sentence_2": string
}
\end{verbatim}

If there are no valid pairs, return:

\begin{verbatim}
{"pairs": []}
\end{verbatim}

Return ONLY valid JSON. No explanation.
\end{promptbox}

\begin{promptbox}{Conflict}
You are given two multi-perspective narratives: Narrative 1 and Narrative 2. Identify all pairs of sentences (one from Narrative 1 and one from Narrative 2) that conflict---that is, both sentences refer to the same specific fact, event, or claim but make clearly opposing or contradictory statements that cannot both be true.

\textbf{Rules:}
\begin{itemize}[leftmargin=*,nosep]
    \item Do NOT include pairs with compatible meanings (those belong to overlap).
    \item Do NOT include pairs that are only loosely related by topic.
    \item The sentences must refer to the same specific fact, event, or claim.
    \item If unsure, exclude the pair.
\end{itemize}

\textbf{Examples of Valid Conflict Pairs:}

\{\{6-shot examples\}\}

\textbf{Examples of Pairs That Look Similar but Are Not Conflict:}

\{\{4 counter-examples\}\}

Now apply the same standard to the narratives below. Before including a pair, always check: (1) do the two sentences refer to the same specific fact, event, or claim---not just the same broad topic, and (2) do they make claims that cannot both be true? If either check fails, exclude the pair.

\textbf{Input:}

Narrative 1: \texttt{\_\_Narrative1\_\_}

Narrative 2: \texttt{\_\_Narrative2\_\_}

Respond with a JSON object with a single key \texttt{"pairs"}, where each item is:

\begin{verbatim}
{
  "sentence_1": string,
  "sentence_2": string
}
\end{verbatim}

If there are no such pairs, return:

\begin{verbatim}
{"pairs": []}
\end{verbatim}

Return ONLY valid JSON. No explanation.
\end{promptbox}

\begin{promptbox}{Unique-A}
You are given two narratives about the same topic, possibly from different perspectives. Identify all sentences from Narrative 1 that are unique relative to Narrative 2---that is, no sentence in Narrative 2 refers to the same specific fact, event, situation, or claim, whether with compatible (overlapping) or incompatible (conflicting) meaning.


Before including any sentence as unique, verify the following:

\begin{enumerate}[leftmargin=*]
    \item Does this sentence appear in any overlap pair between the two narratives?  
    If yes, it has a counterpart in the other narrative. EXCLUDE IT.

    \item Does this sentence appear in any conflict pair between the two narratives?  
    If yes, it has a counterpart in the other narrative. EXCLUDE IT.

    \item Is there any sentence in Narrative 2 that paraphrases or addresses the same specific claim, even with completely different wording?  
    If yes, it has a counterpart. EXCLUDE IT.
\end{enumerate}

Only include the sentence if all three checks pass---the other narrative is completely silent on the specific fact or claim it makes.

\textbf{Common mistakes to avoid:}
\begin{itemize}[leftmargin=*,nosep]
    \item Including a sentence just because its wording does not appear verbatim in the other narrative. Paraphrases count as counterparts.
    \item Including a sentence that expresses a claim the other narrative contradicts. Conflict is still a counterpart relationship.
    \item Including a sentence about a topic the other narrative discusses, even if from a different angle. Check the specific claim, not the topic.
\end{itemize}

\textbf{Examples of Genuinely Unique Sentences:}

\{\{6-shot examples\}\}

\textbf{Examples of Sentences That Look Unique but Are Not:}

\{\{4 counter-examples\}\}

Now apply the same standard to the narratives below. For each sentence in Narrative 1, check every sentence in Narrative 2. Only include the sentence if Narrative 2 is completely silent on the specific fact or claim it makes---no overlap and no conflict.

\textbf{Input:}

Narrative 1: \texttt{\_\_Narrative1\_\_}

Narrative 2: \texttt{\_\_Narrative2\_\_}

Respond with a JSON object with a single key \texttt{"items"} whose value is an array of objects. EVERY object MUST include both keys:

\texttt{"sentence\_1"}: string (the unique sentence from Narrative 1).

\texttt{"sentence\_2"}: always the empty string \texttt{""}.

Example:

\begin{verbatim}
{"items":[{"sentence_1":string,
"sentence_2":""}]}
\end{verbatim}

If there are no unique sentences, return:

\begin{verbatim}
{"items": []}
\end{verbatim}

Return ONLY valid JSON. No markdown. No explanation. Use \texttt{"items"}, not \texttt{"pairs"}.
\end{promptbox}

\begin{promptbox}{Unique-B}
You are given two narratives about the same topic, possibly from different perspectives. Identify all sentences from Narrative 2 that are unique relative to Narrative 1---that is, no sentence in Narrative 1 refers to the same specific fact, event, situation, or claim, whether with compatible (overlapping) or incompatible (conflicting) meaning.


\textbf{IMPORTANT: CROSS-CHECK BEFORE INCLUDING:}

Before including any sentence as unique, verify the following:

\begin{enumerate}[leftmargin=*]
    \item Does this sentence appear in any overlap pair between the two narratives?  
    If yes, it has a counterpart in the other narrative. EXCLUDE IT.

    \item Does this sentence appear in any conflict pair between the two narratives?  
    If yes, it has a counterpart in the other narrative. EXCLUDE IT.

    \item Is there any sentence in Narrative 2 that paraphrases or addresses the same specific claim, even with completely different wording?  
    If yes, it has a counterpart. EXCLUDE IT.
\end{enumerate}

Only include the sentence if all three checks pass---the other narrative is completely silent on the specific fact or claim it makes.

\textbf{Common mistakes to avoid:}
\begin{itemize}[leftmargin=*,nosep]
    \item Including a sentence just because its wording does not appear verbatim in the other narrative. Paraphrases count as counterparts.
    \item Including a sentence that expresses a claim the other narrative contradicts. Conflict is still a counterpart relationship.
    \item Including a sentence about a topic the other narrative discusses, even if from a different angle. Check the specific claim, not the topic.
\end{itemize}

\textbf{Examples of Genuinely Unique Sentences:}

\{\{5-shot examples\}\}

\textbf{Examples of Sentences That Look Unique but Are Not:}

\{\{4 counter-examples\}\}

Now apply the same standard to the narratives below. For each sentence in Narrative 2, check every sentence in Narrative 1. Only include the sentence if Narrative 1 is completely silent on the specific fact or claim it makes---no overlap and no conflict.

\textbf{Input:}

Narrative 1: \texttt{\_\_Narrative1\_\_}

Narrative 2: \texttt{\_\_Narrative2\_\_}

Respond with a JSON object with a single key \texttt{"items"} whose value is an array of objects. EVERY object MUST include both keys:

\texttt{"sentence\_1"}: always the empty string \texttt{""}.

\texttt{"sentence\_2"}: string (the unique sentence from Narrative 2).

Example:

\begin{verbatim}
{"items":[{"sentence_1":"",
"sentence_2":string}]}
\end{verbatim}

If there are no unique sentences, return:

\begin{verbatim}
{"items": []}
\end{verbatim}

Return ONLY valid JSON. No explanation.
\end{promptbox}

\subsubsection{Chain-of-Thought Prompt}

\begin{promptbox}{Overlap}
You are given two multi-perspective narratives: Narrative 1 and Narrative 2. Identify all pairs of sentences (one from Narrative 1 and one from Narrative 2) that overlap---that is, both sentences describe the same specific fact, event, situation, or claim and express compatible meaning (both can be true at the same time).

\textbf{Rules:}
\begin{itemize}[leftmargin=*,nosep]
    \item Do NOT include pairs with incompatible or contradictory meanings.
    \item Do NOT match sentences based only on broad topic similarity.
    \item The sentences must refer to the same specific fact, event, or claim.
    \item Do NOT include pairs with weak or vague similarities.
    \item If unsure, exclude the pair.
\end{itemize}

\textbf{Input:}

Narrative 1: \texttt{\_\_Narrative1\_\_}

Narrative 2: \texttt{\_\_Narrative2\_\_}

Before producing the final JSON, reason through the following steps:

\textbf{Step 1}: List every sentence from Narrative 1, numbered.

\textbf{Step 2}: List every sentence from Narrative 2, numbered.

\textbf{Step 3}: Identify all aspects, facts, events, or claims that appear in BOTH narratives (not just the same general topic; the same specific target must be addressed in both).

\textbf{Step 4}: For each common aspect identified in Step 3, find the sentence(s) from Narrative 1 and Narrative 2 that address it and form candidate pairs.

\textbf{Step 5}: For each candidate pair, decide: are the meanings compatible
toward that aspect (both can be true at the same time)?

If yes $\rightarrow$ overlap.

If the meanings contradict $\rightarrow$ exclude (that is conflict, not overlap).

If the match is vague or topical only $\rightarrow$ exclude.

\textbf{Step 6} --- Output only the confirmed overlap pairs as JSON.

Respond with your reasoning first, then end with a single JSON object:

\begin{verbatim}
{
  "pairs": [{"sentence_1": string, 
  "sentence_2": string}]
}
\end{verbatim}

If there are no valid pairs, end with:

\begin{verbatim}
{"pairs": []}
\end{verbatim}

Return ONLY your reasoning followed by valid JSON. 
\end{promptbox}

\begin{promptbox}{Conflict}
You are given two narratives: Narrative 1 and Narrative 2. Each narrative consists of multiple sentences. Identify all pairs of sentences (one from Narrative 1 and one from Narrative 2) that conflict---that is, both sentences refer to the same specific fact, event, or claim but make clearly opposing or contradictory statements that cannot both be true.

\textbf{Rules:}
\begin{itemize}[leftmargin=*,nosep]
    \item Do NOT include pairs with compatible meanings (those belong to overlap).
    \item Do NOT include pairs that are only loosely related by topic.
    \item The sentences must refer to the same specific fact, event, or claim.
    \item If unsure, exclude the pair.
\end{itemize}

\textbf{Input:}

Narrative 1: \texttt{\_\_Narrative1\_\_}

Narrative 2: \texttt{\_\_Narrative2\_\_}

Before producing the final JSON, reason through the following steps:

\textbf{Step 1}: List every sentence from Narrative 1, numbered.

\textbf{Step 2}: List every sentence from Narrative 2, numbered.

\textbf{Step 3}: Identify all aspects, facts, events, or claims that appear in BOTH narratives (not just the same general topic; the same specific target must be addressed in both).

\textbf{Step 4}: For each common aspect identified in Step 3, find the sentence(s) from Narrative 1 and Narrative 2 that address it and form candidate pairs.

\textbf{Step 5}: For each candidate pair, decide: do the two sentences take opposing or contradictory positions toward that aspect such that both cannot be true at the same time?

If yes $\rightarrow$ conflict.

If the meanings are compatible $\rightarrow$ exclude (that is overlap, not conflict).

If the match is vague or topical only $\rightarrow$ exclude.

\textbf{Step 6} --- Output only the confirmed conflict pairs as JSON.

Respond with your reasoning first, then end with a single JSON object:

\begin{verbatim}
{
  "pairs": [{"sentence_1": string, 
  "sentence_2": string}]
}
\end{verbatim}

If there are no such pairs, end with:

\begin{verbatim}
{"pairs": []}
\end{verbatim}

Return ONLY your reasoning followed by valid JSON.
\end{promptbox}

\begin{promptbox}{Unique-A}
You are given two narratives about the same topic, possibly from different perspectives. Identify all sentences from Narrative 1 that are unique relative to Narrative 2---that is, no sentence in Narrative 2 refers to the same specific fact, event, situation, or claim, whether with compatible (overlapping) or incompatible (conflicting) meaning.

\textbf{Input:}

Narrative 1: \texttt{\_\_Narrative1\_\_}

Narrative 2: \texttt{\_\_Narrative2\_\_}

Before producing the final JSON, reason through the following steps:

\textbf{Step 1}: List every sentence from Narrative 1, numbered.

\textbf{Step 2}: List every sentence from Narrative 2, numbered.

\textbf{Step 3}: Identify all aspects, facts, events, or claims that appear
in BOTH narratives (the shared aspects).

\textbf{Step 4}: For each sentence in Narrative 1, check whether its aspect
or target appears in the shared aspects identified in Step 3.

If yes $\rightarrow$ it has a counterpart (overlap or conflict); exclude.

If no $\rightarrow$ it is a candidate for unique.

\textbf{Step 5} --- Output only the confirmed unique sentences as JSON.

Respond with your reasoning first, then end with a single JSON object where EVERY object includes both keys:

\texttt{"sentence\_1"}: string (the unique sentence from Narrative 1).

\texttt{"sentence\_2"}: always the empty string \texttt{""}.

\begin{verbatim}
{"items": [{"sentence_1": string, 
"sentence_2": ""}]}
\end{verbatim}

If there are no unique sentences, end with:

\begin{verbatim}
{"items": []}
\end{verbatim}

Return ONLY your reasoning followed by valid JSON. 
\end{promptbox}

\begin{promptbox}{Unique-B}
You are given two narratives about the same topic, possibly from different perspectives. Identify all sentences from Narrative 2 that are unique relative to Narrative 1---that is, no sentence in Narrative 1 refers to the same specific fact, event, situation, or claim, whether with compatible (overlapping) or incompatible (conflicting) meaning.

\textbf{Input:}

Narrative 1: \texttt{\_\_Narrative1\_\_}

Narrative 2: \texttt{\_\_Narrative2\_\_}

Before producing the final JSON, reason through the following steps:

\textbf{Step 1}: List every sentence from Narrative 1, numbered.

\textbf{Step 2}: List every sentence from Narrative 2, numbered.

\textbf{Step 3}: Identify all aspects, facts, events, or claims that appear
in BOTH narratives (the shared aspects).

\textbf{Step 4}: For each sentence in Narrative 2, check whether its aspect
or target appears in the shared aspects identified in Step 3.

If yes $\rightarrow$ it has a counterpart (overlap or conflict); exclude.

If no $\rightarrow$ it is a candidate for unique.

\textbf{Step 5}: Output only the confirmed unique sentences as JSON.

Respond with your reasoning first, then end with a single JSON object
where EVERY object includes both keys:

\texttt{"sentence\_1"}: always the empty string \texttt{""}.

\texttt{"sentence\_2"}: string (the unique sentence from Narrative 2).

\begin{verbatim}
{"items": [{"sentence_1": "", 
"sentence_2": string}]}
\end{verbatim}

If there are no unique sentences, end with:

\begin{verbatim}
{"items": []}
\end{verbatim}

Return ONLY your reasoning followed by valid JSON.
\end{promptbox}

\subsubsection{Combined Prompt.}

\begin{promptbox}{User Prompt}
You are an extractor who compares multi-perspective narratives
about the same topic. The task is to extract four categories of information from the narratives.

1. OVERLAP: a sentence pair (one from each narrative) about the
same target where neither sentence asserts something the other denies, and for evaluative statements, both take the same stance (positive, or negative, or neutral) toward that target.

2. CONFLICT:  a sentence pair about the same target that is
factually incompatible (different numbers, causal attribution,
actor, timing, confirm vs.\ deny), OR an evaluative pair that
takes opposing stances toward the same target.

3. UNIQUE TO NARRATIVE 1: a sentence in Narrative 1 whose target
is not addressed, confirmed, contradicted, or evaluated by any
sentence in Narrative 2.

4. UNIQUE TO NARRATIVE 2: a sentence in Narrative 2 whose target
is not addressed, confirmed, contradicted, or evaluated by any
sentence in Narrative 2.

\textbf{Rules:}
\begin{itemize}[leftmargin=*,nosep]
    \item Copy sentences verbatim; do not paraphrase.
    \item Only pair sentences with the same target, not just the same
    general subject.
    \item Overlap and conflict use paired sentences; unique entries use
    items with one empty side.
    \item A sentence may appear in more than one pair only if it covers
    more than one distinct target, each matched to a different
    counterpart. Do not pair a sentence with multiple counterparts
    just to increase coverage.
    \item Return ONLY valid JSON. No explanation.
\end{itemize}

\textbf{Input:}

\textbf{Narrative 1:} \texttt{\_\_Narrative1\_\_}

\textbf{Narrative 2:} \texttt{\_\_Narrative2\_\_}

Respond with a single JSON object:

\begin{verbatim}
{
  "overlap": {
    "pairs": [
      {"sentence_1": string, 
      "sentence_2": string}
    ]
  },
  "conflict": {
    "pairs": [
      {"sentence_1": string, 
      "sentence_2": string}
    ]
  },
  "unique1": {
    "items": [
      {"sentence_1": string, 
      "sentence_2": ""}
    ]
  },
  "unique2": {
    "items": [
      {"sentence_1": "", 
      "sentence_2": string}
    ]
  }
}
\end{verbatim}
Return ONLY valid JSON.
\end{promptbox}

\subsubsection{Bottleneck Analysis Prompt}
\label{app:prompt-bottleneck}

\begin{promptbox}{Relation Understanding (Overlap)}
You are a strict sentence-pair classifier. Your task is to decide whether a candidate sentence pair is a valid OVERLAP pair.

\textbf{Definition of OVERLAP:}

A pair is OVERLAP only if \texttt{sentence\_1} and \texttt{sentence\_2} refer to the same specific fact, event, claim, attribute, or aspect, and express compatible meaning.

This means:
\begin{itemize}[leftmargin=*,nosep]
    \item The two sentences describe the same underlying information.
    \item The two sentences can both be true at the same time.
    \item The second sentence either repeats, paraphrases, confirms, or gives a compatible version of the information in the first sentence.
    \item Minor wording differences are allowed.
    \item Minor differences in specificity are allowed if the core fact remains the same.
\end{itemize}

A pair is NOT OVERLAP if:
\begin{itemize}[leftmargin=*,nosep]
    \item The sentences are only about the same broad topic or event, but discuss different facts.
    \item The sentences discuss different aspects, such as one sentence giving a location and the other giving a casualty count.
    \item The sentences make incompatible or contradictory claims.
    \item The sentences are related by background context but do not express the same claim.
    \item One sentence is more general but does not clearly support the same specific fact.
    \item The pair requires guessing, external knowledge, or loose inference to connect them.
\end{itemize}

\textbf{Labeling rule:}
\begin{itemize}[leftmargin=*,nosep]
    \item Choose \texttt{"yes"} only for a valid overlap pair.
    \item Choose \texttt{"no"} for conflict pairs, unique/unmatched information, wrong-facet pairs, loosely related pairs, or unrelated pairs.
\end{itemize}


\textbf{Input:}

Narrative 1:

\texttt{\_\_NARRATIVE1\_\_}

Narrative 2:

\texttt{\_\_NARRATIVE2\_\_}

Candidate pair:

\texttt{sentence\_1: \_\_SENTENCE1\_\_}

\texttt{sentence\_2: \_\_SENTENCE2\_\_}

\textbf{Question:}

Is this candidate pair a valid OVERLAP pair?

Respond with JSON only:

\begin{verbatim}
{"label": "yes"}
\end{verbatim}
\end{promptbox}

\begin{promptbox}{Relation Understanding (Conflict)}
You are a strict sentence-pair classifier. Your task is to decide whether a candidate sentence pair is a valid CONFLICT pair.

\textbf{Definition of CONFLICT:}

A pair is CONFLICT only if \texttt{sentence\_1} and \texttt{sentence\_2} refer to the same specific fact, event, claim, attribute, or aspect, but express incompatible meanings that cannot both be true at the same time.

This means:
\begin{itemize}[leftmargin=*,nosep]
    \item The two sentences must discuss the same underlying information and must make clearly opposing or mutually inconsistent claims.
    \item The disagreement must be about the same target, aspect, value, status, cause, time, location, person, number, responsibility, or event detail.
    \item The contradiction may involve polarity, numeric values, dates/times, locations, identities, causal explanations, responsibility, status, or reported outcomes.
    \item Minor wording differences are allowed only if the core disagreement remains clear.
\end{itemize}

A pair is NOT CONFLICT if:
\begin{itemize}[leftmargin=*,nosep]
    \item The sentences are only about the same broad topic or event but discuss different facts.
    \item The sentences discuss different aspects, such as one sentence giving a location and the other giving a casualty count.
    \item The sentences express compatible meaning or can both be true at the same time.
    \item One sentence gives additional detail that is not contradicted by the other.
    \item The pair is an overlap pair, a unique/unmatched pair, a wrong-facet pair, a loosely related pair, or an unrelated pair.
\end{itemize}

\textbf{Labeling rule:}
\begin{itemize}[leftmargin=*,nosep]
    \item Choose \texttt{"yes"} only for a valid conflict pair.
    \item Choose \texttt{"no"} for overlap pairs, unique/unmatched information, wrong-facet pairs, loosely related pairs, or unrelated pairs.
\end{itemize}


\textbf{Input:}

Narrative 1:

\texttt{\_\_NARRATIVE1\_\_}

Narrative 2:

\texttt{\_\_NARRATIVE2\_\_}

Candidate pair:

\texttt{sentence\_1: \_\_SENTENCE1\_\_}

\texttt{sentence\_2: \_\_SENTENCE2\_\_}

\textbf{Question:}

Is this candidate pair a valid CONFLICT pair?

Respond with JSON only:

\begin{verbatim}
{"label": "yes"}
\end{verbatim}
\end{promptbox}


\begin{promptbox}{Pair Alignment (Overlap)}
You are a strict sentence-pair aligner. Your task is to identify valid OVERLAP pairs from two candidate sentence lists
(Narrative 1 candidates and Narrative 2 candidates). You are given only the candidate sentences.

\textbf{Definition of OVERLAP:}

A pair is OVERLAP only if \texttt{sentence\_1} and \texttt{sentence\_2} refer to the same specific fact, event, claim, attribute, or aspect, and express compatible meaning.

This means:
\begin{itemize}[leftmargin=*,nosep]
    \item The two sentences describe the same underlying information.
    \item The two sentences can both be true at the same time.
    \item The second sentence either repeats, paraphrases, confirms, or gives a compatible version of the information in the first sentence.
    \item Minor wording differences are allowed.
    \item Minor differences in specificity are allowed if the core fact remains the same.
\end{itemize}

A pair is NOT OVERLAP if:
\begin{itemize}[leftmargin=*,nosep]
    \item The sentences are only about the same broad topic or event but discuss different facts.
    \item The sentences discuss different aspects, such as one sentence giving a location and the other giving a casualty count.
    \item The sentences make incompatible or contradictory claims.
    \item One sentence gives information that is absent from the other.
    \item The sentences are related by background context but do not express the same claim.
    \item One sentence is more general but does not clearly support the same specific fact.
    \item The pair requires guessing, external knowledge, or loose inference to connect them.
\end{itemize}

\textbf{Alignment rules:}
\begin{itemize}[leftmargin=*,nosep]
    \item \texttt{sentence\_1} must be chosen from the Narrative 1 candidate list.
    \item \texttt{sentence\_2} must be chosen from the Narrative 2 candidate list.
    \item Return only valid overlap pairs.
\end{itemize}

\textbf{Input}

Narrative 1 candidate sentences:

\texttt{\_\_CANDIDATES1\_\_}

Narrative 2 candidate sentences:

\texttt{\_\_CANDIDATES2\_\_}

Respond with JSON only:

\begin{verbatim}
{"pairs": [{"sentence_1": "...", 
"sentence_2": "..."}]}
\end{verbatim}

\end{promptbox}

\begin{promptbox}{Pair Alignment (Conflict)}
You are a strict sentence-pair aligner. Your task is to identify valid CONFLICT pairs from two candidate sentence lists
(Narrative 1 candidates and Narrative 2 candidates). You are given only the candidate sentences.

\textbf{Definition of CONFLICT:}

A pair is CONFLICT only if \texttt{sentence\_1} and \texttt{sentence\_2} refer to the same specific fact, event, claim, attribute, or aspect, but express incompatible meanings that cannot both be true at the same time.

This means:
\begin{itemize}[leftmargin=*,nosep]
    \item The two sentences must discuss the same underlying information.
    \item The disagreement must be about the same target, aspect, value, status, cause, time, location, person, number, responsibility, or event detail.
    \item The two sentences must make clearly opposing or mutually inconsistent claims.
    \item The contradiction may involve polarity, numeric values, dates/times, locations, identities, causal explanations, responsibility, status, or reported outcomes.
    \item Minor wording differences are allowed only if the core disagreement remains clear.
\end{itemize}

A pair is NOT CONFLICT if:
\begin{itemize}[leftmargin=*,nosep]
    \item The sentences are only about the same broad topic or event but discuss different facts.
    \item The sentences discuss different aspects, such as one sentence giving a location and the other giving a casualty count.
    \item The sentences express compatible meaning or can both be true at the same time.
    \item One sentence gives additional detail that is not contradicted by the other.
    \item One sentence gives information that is absent from the other.
    \item The sentences are merely different, incomplete, or unequal in specificity.
\end{itemize}

\textbf{Alignment rules:}
\begin{itemize}[leftmargin=*,nosep]
    \item \texttt{sentence\_1} must be chosen from the Narrative 1 candidate list.
    \item \texttt{sentence\_2} must be chosen from the Narrative 2 candidate list.
    \item Return only valid conflict pairs.
\end{itemize}

\textbf{Input}

Narrative 1 candidate sentences:

\texttt{\_\_CANDIDATES1\_\_}

Narrative 2 candidate sentences:

\texttt{\_\_CANDIDATES2\_\_}

Respond with JSON only:

\begin{verbatim}
{"pairs": [{"sentence_1": "...", 
"sentence_2": "..."}]}
\end{verbatim}

\end{promptbox}


\newpage
\subsection{Data Samples}
\label{app:data-samples}
\scriptsize

\begin{longtable}{
    >{\justifying\arraybackslash}p{0.47\textwidth}
    >{\justifying\arraybackslash}p{0.47\textwidth}
}
\caption{Examples of Overlap, Conflict, and Unique information across
multi-perspective narrative pairs. Each row corresponds to one pair of narratives.
Matching \ov{$n$} identifiers denote sentence pairs that express overlapping
information, while matching \cf{$n$} identifiers denote sentence pairs that
express conflicting information. Sentences marked with \ua{$n$} and \ub{$n$}
contain information unique to Narrative A and Narrative B, respectively.}
\label{tab:data-examples}\\

\hline
\textbf{Narrative A} & \textbf{Narrative B} \\
\hline
\endfirsthead

\multicolumn{2}{c}%
{{\tablename\ \thetable{} -- continued from previous page}}\\
\midrule
\textbf{Narrative A} & \textbf{Narrative B} \\
\midrule
\endhead

\hline
\multicolumn{2}{r}{\textit{Continued on next page}}\\
\endfoot

\hline
\endlastfoot

\addlinespace[1pt]
\multicolumn{2}{>{\columncolor{domainbg}}c}{\textbf{Political News}}\\
\midrule

\ov{1} A federal judge is striking down the Trump administration's highly touted small-business health insurance plan, calling it an ``end run'' around consumer protections.
\ov{2} The ruling Thursday by U.S. District Judge John Bates in Washington, D.C., is the second setback in a week for the administration's healthcare initiatives.
\cf{1} On Wednesday another federal judge blocked Medicaid work requirements for low-income people.
\ov{3} At issue in the latest ruling are so-called ``association health plans,'' in which businesses and sole proprietors can band together to offer lower-cost coverage that doesn't provide all the benefits required under the Affordable Care Act.
\ov{4} President Donald Trump has hailed the small-business plans as a big success, but their impact is difficult to measure.
\ua{1} Unable to repeal Obamacare, as ACA is known, in Congress, the Trump administration has tried to use its rule-making powers to create room for alternatives.

&

\ov{1} A federal court in Washington nullified the Trump administration's touted small-business health-insurance scheme, describing it as an attempt to sidestep established consumer safeguards.
\cf{1} The Wednesday decision actually upheld the Medicaid work-requirement rule, permitting states to enforce employment conditions on low-income beneficiaries.
\ov{2} The Thursday decision added to a series of recent legal challenges the administration has faced this week.
\ub{1} Some analysts suggest that the administration may have already accomplished its objectives regarding the ACA, potentially reducing the need for additional rule-making.
\ov{4} President Trump has touted the small-business insurance options as a major triumph, even though their real effect remains hard to quantify.
\ov{3} The controversy centers on so-called association health plans, which let small firms and independent contractors combine to purchase cheaper policies that omit several ACA-mandated benefits.
\ub{2} Some Republican leaders have praised the administration's effort to increase market competition in health care, suggesting it could help lower premiums.
\ub{3} Democratic health-care advocates have expressed concern that changes to ACA protections could affect vulnerable patients, urging congressional attention.
\\[2mm]
\midrule
\midrule

\ov{1} When throngs of abortion protesters convene Friday for the annual March for Life, they will enjoy the warmest show of White House support in the event's four-decade history.
\ov{2} Vice President Mike Pence and top Trump adviser Kellyanne Conway will headline the pre-march rally on the National Mall, becoming the highest-ranking members of a presidential administration ever to speak at the annual protest.
\cf{1} President Trump plans to call in to the rally to voice his support, following a pattern set by two of his GOP predecessors.
\ov{3} Pence and Conway's presence is a huge boon to the march, attended by hundreds of thousands of people every year who want to register their opposition to the Supreme Court's 1973 Roe v. Wade decision legalizing abortion.
\ua{1} It further solidifies their confidence that President Trump will live up to his promises to crack down on abortion, as he vowed during his campaign.
\ov{4} Pence is a favorite of conservatives for his actions as Indiana governor and as a member of Congress to limit abortion.
\ua{2} While Trump used to support abortion rights, activists view Pence as one of their own, a longtime and trustworthy ally.

&

\ov{1} Friday's March for Life will see an unprecedented level of backing from the White House in its 40-year history.
\ov{2} Vice President Mike Pence and senior Trump aide Kellyanne Conway are slated to address the pre-march gathering on the National Mall, marking the first time officials of such rank have spoken at the protest.
\cf{1} President Trump has declined to join the event remotely, breaking with the precedent set by his Republican predecessors.
\ov{3} The appearance of the administration's top figures is expected to energize the crowd, which regularly draws massive numbers of anti-abortion demonstrators.
\ub{1} Critics argue the administration's involvement does little to assure that any meaningful restrictions on abortion will be enacted.
\ov{4} Conservatives have long praised Pence for his record on abortion restrictions during his tenure as Indiana's governor and as a congressman.
\ub{2} Pence is also scheduled to meet with senior staff from the Department of Health and Human Services on Friday to discuss a new maternal-health initiative, a gathering separate from the March for Life.
\ub{3} Lawmakers from both parties are expected to debate a new health-care bill later this week, a measure that does not address abortion policy directly.
\\[2mm]

\midrule
\midrule

\ov{1} President-elect Donald Trump and intelligence officials appear to be at odds over a briefing about Russia's interference in the 2016 election.
\cf{1} Trump says that briefing was delayed.
\ov{2} The incoming president took to Twitter to share his opinion.
\ov{3} ``The `intelligence' briefing on so-called `Russian-hacking' was delayed until Friday. Perhaps more time needed to build a case. Very strange!'' Trump tweeted.
\cf{2} Meanwhile, intelligence officials say the meeting was scheduled for later in the week and that President Barack Obama hasn't even received the full briefing yet.
\ov{4} This latest dust up represents a growing divide between the president-elect and intelligence agencies.
\ov{5} The U.S. intelligence community has been investigating Russia's ties to election-related hacking of the DNC.
\cf{3} But Trump has questioned whether or not Russia was even involved.
\ua{1} ``Julian Assange said a 14-year-old could have hacked Podesta -- why was DNC so careless?'' Trump questioned in a Tweet.

&

\ov{1} President-elect Donald Trump finds himself in disagreement with members of the intelligence community regarding the timing of a briefing on alleged Russian election meddling.
\ov{2} He used his Twitter account to voice his thoughts on the matter.
\ov{3} In a post, Trump suggested the postponement might be due to the need for a stronger evidentiary foundation, calling the situation odd.
\ov{5} Federal intelligence agencies continue to probe alleged connections between Moscow and the breach of Democratic National Committee servers.
\cf{1} Agency representatives insist the schedule was set as originally planned, refuting Trump's suggestion of a delay.
\cf{2} Senior officials have indicated that the comprehensive briefing had already been handed over to the outgoing administration before the weekend.
\cf{3} Intelligence analysts continue to assert that Russian actors played a direct role in the cyber intrusions.
\ov{4} The recent spat underscores an expanding rift between the incoming administration and the nation's intelligence services.
\ub{1} Critics note that the intelligence community is also monitoring other cyber threats, including activities attributed to China.
\ub{2} The inauguration ceremony is slated for January 20, with preparations already underway at the Capitol.
\tabularnewline[2mm]
\hline

\addlinespace[1pt]
\multicolumn{2}{>{\columncolor{domainbg}}c}{\textbf{Factual News}}\\
\midrule

\ov{1} A man stormed into a Zurich mosque on Monday evening and opened fire on people praying, injuring three, Swiss police said.
\ov{2} They said they had collected evidence inside the building and would make more details available on Tuesday.
\ov{3} Two of the three men aged 30, 35 and 56, were seriously injured in the attack shortly after 5:30 p.m. local time (1630 GMT) near the main train station in Switzerland's financial capital, Zurich police said.
\ov{4} A third sustained less severe injuries. All three were brought to hospital.
\cf{1} The unidentified suspect was described as a man around 30 years old.
\ov{5} Witnesses said he was wearing dark clothing and a dark wool cap.
\ua{1} He fled the mosque, police said.
\ov{6} Police later confirmed that the firearm used in the attack had been legally owned and the shooter possessed a valid gun permit.
\ov{7} Police said a body was found nearby but would not comment on any link to the shootings while investigations continued.
\ov{8} People at the scene told Reuters the Islamic Center on Zurich's Eisgasse was used as a mosque, often by Somalis.

&

\cf{1} A gunman who shot three worshippers in a Zurich mosque on Monday evening was a 24-year-old Swiss man, police said on Tuesday.
\ub{1} He was reported to have Ghanaian roots and no apparent links to Islamist radicalism.
\ub{2} He seems to have taken his own life shortly after the mosque shooting, whose motivation remains a mystery, police officials said.
\ov{1} The gunman, from the nearby town of Uster, had stormed into the Islamic centre near the main train station in Switzerland's financial capital and opened fire on people praying, wounding three men, whose condition was said to be improving on Tuesday.
\ov{7} His body was found soon afterwards about 300 metres away.
\ov{6} The mayhem continued when he entered the mosque after dusk on Monday, armed with a gun for which he had a permit.
\ov{3} The victims, aged 30, 35 and 56, included two who were seriously injured and a third with less severe injuries.
\ov{5} Witnesses described the attacker as wearing dark clothing and a dark wool cap as he entered the mosque.
\ov{8} The Islamic centre on Eisgasse is known to be frequented by Somali worshippers, and the three injured men were identified as Somalis.
\ov{4} All three victims were transported to a nearby hospital for treatment.
\ov{2} Police also gathered forensic evidence inside the mosque after the shooting.
\\[2mm]
\midrule
\midrule

\ov{1} ``This shelter mission is going to be a very heavy lift. We're anticipating over 30,000 people being placed in shelters temporarily.''
\ov{2} Already 5,500 people are staying in shelters in Houston, and Turner only expects that number to rise in a greater metro area of some 6.5 million people.
\ov{3} Texas Gov. Greg Abbott activated the state's entire National Guard on Monday, saying roughly 12,000 guard members will be deployed to respond to Harvey.
\cf{1} Initially, about 3,000 national and state guard members had been deployed.
\ua{1} Eighteen of those counties have been granted a federal disaster declaration, a move that triggers FEMA's support.
\ov{4} The U.S. Coast Guard has also been active in the response, deploying at least eight helicopters and requesting 11 more from across the country to conduct rescues.
\ov{5} All told, Pence said Monday there are more than 8,500 federal personnel on the ground in the area to help state and local officials, who are leading the rescue efforts.
\cf{2} Officials have reported no confirmed flood-related fatalities to date.
\cf{3} The storm has dumped over 9 trillion gallons of water across the region, according to state estimates.
\ov{6} ``This is going to take us months, years to get back to normal,'' Emmett said.

&

\ov{2} The federal emergency agency said earlier today that more than 5,500 people already are in shelters.
\cf{2} Authorities have confirmed at least six flood-related deaths in Houston and another in coastal Rockport.
\ov{1} It expects that number to reach 30,000 eventually.
\ov{3} He noted that the entire 12,000-strong Texas National Guard has been deployed.
\cf{1} Initially, only about 1,000 guard members were dispatched before the full deployment.
\ub{1} The governor made a state disaster declaration for 54 counties.
\ov{6} FEMA administrator Brock Long put the disaster into perspective with a single line during the presser: ``We are going to be here for several years helping you guys recover.''
\ov{4} The Coast Guard has deployed eight helicopters to assist with air rescues, with additional aircraft requested.
\ov{5} Over 8,500 federal responders have been stationed in the affected region to support local officials.
\cf{3} The rain continues to fall ... some 11 trillion gallons of water already has fallen in the area.
\tabularnewline[2mm]
\hline

\addlinespace[1pt]
\multicolumn{2}{>{\columncolor{domainbg}}c}{\textbf{Peer Review}}\\
\midrule

\ov{1} This paper addresses the generalization of adversarial training by proposing a new domain adaptation method.
\ov{2} The proposed technique offers a modest novelty compared to existing domain-adaptation adversarial methods.
\ov{3} In order to have robust defense for adversarial examples, they combine supervised and unsupervised learning for domain adaptation.
\ua{1} The idea of domain adaptation is to increase the similarity between clear and adversarial examples.
\ua{2} For this purpose, in their objective, they are minimizing the domain shift by aligning the covariance matrix and mean vector of the clean and adversarial examples.
\cf{1} From experimental viewpoint, they have lower performance than almost all competitors on clean data.
\ov{4} However, they are beating most of the competitors under white-box threats.
\ov{5} And also have better results under black-box threats.
\ov{6} So that means their method gives a good generalization against different attacks.
\cf{2} In CIFAR-100 they do not have this trade-off for accuracy and generalization; they are beating other competitors in clean data as well.
\ov{7} The paper is clear and well-written.
\ov{8} The introduction and background give useful information.
\ua{3} In general, I think the paper has a potential for acceptance, but I have to mention that I am not an expert in Adversarial networks area.
\ov{9} The experimental evidence remains inconclusive, and additional ablation studies would strengthen the claims.
\cf{3} The approach incurs a substantial computational overhead due to the covariance alignment step.
\cf{4} The authors compare against a wide range of strong baselines, ensuring a fair evaluation.

&

\ov{1} Authors propose a new adversarial training with domain adaptation method to overcome the weak generalisation problem in adversarial training for adversarial examples from different attacks.
\ov{8} The introduction provides a concise overview of related domain-adaptation techniques, which is helpful for readers.
\ub{1} Authors consider the adversarial training as a domain adaptation task with limited number of target labeled data.
\ov{3} They demonstrate that by combining unsupervised and supervised domain adaptation with adversarial training, the generalisation ability on adversarial examples from various attacks can be improved for efficient defence.
\ov{4} The experimental results on several benchmark datasets show better performance under white-box attacks compared to current competing methods.
\ov{5} The method also performs better under black-box attacks.
\ov{6} Overall, their results show better generalisation against different adversarial attacks in most cases.
\cf{2} Nevertheless, on CIFAR-100 a noticeable trade-off between clean accuracy and robustness remains, limiting its advantage.
\ov{7} Paper is clearly written and well structured.
\cf{3} The algorithm runs with comparable runtime to standard adversarial training, adding negligible overhead.
\ov{2} The novelty of the proposed technique is fair and the originality alike.
\cf{4} However, the set of baselines is limited and omits several recent strong defenses, weakening the evaluation.
\ov{9} The results are not very conclusive therefore I think more experiments are needed and possible further adjustments.
\cf{1} Across all benchmarks the method attains the highest clean-image accuracy among the compared approaches.\\[2mm]
\midrule
\midrule

\ua{1} Vanilla Q learning often suffers from the over estimation problem i.e, it might choose bad actions initially which gets reinforced later on.
\ua{2} Double Q learning maintains two Q functions $Q^A$ and $Q^B$ which are used to update each other.
\ov{1} By maintaining two separate estimators, Double Q-learning reduces the overestimation bias inherent in vanilla Q-learning.
\ua{3} This helps in the case of random rewards.
\cf{1} However, the complicated coupled dynamics of $Q^A$ and $Q^B$ makes the analysis much harder.
\ua{4} Even though there are many satisfactory analyses for vanilla Q learning, double Q learning is lacking in this front.
\ov{2}\ov{3}\cf{2} This paper attempts to fill in this gap by giving a non-asymptotic convergence analysis of DQL, which is much better than the previous analyses in terms of every parameter.
\ov{4} In both the synchronous and asynchronous case (i.e all cases updated at once or updated from a stream of off-policy data respectively) they obtain a time complexity of $O(1/\epsilon^2(1-\gamma)^7)$ where $\epsilon$ is the target error and $\gamma$ is the discount factor.
\ov{5} The proof technique mentioned in Section 3.3 considers the error propagation dynamics and analyzes the iterations using standard concentration inequalities.
\cf{3} The convergence proof crucially relies on a decaying step-size schedule, not a constant one.
\ov{6} I think this is an important problem with lots of recent interest.
\ov{7} The paper is fairly well written and makes a good attempt at obtaining a finite time analysis of DQL.
\ov{8} As mentioned in the paper, the results are worse than the guarantees obtained for the vanilla Q learning in terms of the discount factor $\gamma$ which is the major drawback of the work.
\ua{5} The paper does not provide a conclusive empirical comparison with vanilla Q-learning.
\ov{9} All in all, I expect that the techniques established here would be useful in further investigation of DQL and eventually show that it outperforms vanilla Q learning.

&

\ov{2} The paper studied double Q-learning in RL and established finite time convergence analysis for both synchronous/asynchronous version of the algorithm.
\ov{3} The complexity result improves the existing result in terms of the dependency on parameters such as $\epsilon, 1-\gamma, \cdots$.
\cf{2} The improvement over prior work is modest, affecting only a subset of parameters.
\ub{1} This papers studied Double Q-learning, which is one of the most widely used technique in RL.
\ov{1} Double Q-learning mitigates the overestimation bias that plagues vanilla Q-learning.
\ov{2} The authors established the non-asymptotic convergence result of both synchronous/asynchronous Double Q-learning for tabular discounted MDP.
\ov{4} In particular, the paper proved $O(\frac{1}{(1-\gamma)^7\epsilon^2})$ time complexity for both methods (ignoring other constants), improving upon the existing result that adopts a decaying step size.
\cf{3} This is achieved by using more aggressive constants step size and a more refined analysis.
\ov{5} The analysis relies on standard concentration inequalities to control the error propagation.
\cf{1} Despite the coupled dynamics, the analysis proceeds without significant technical hurdles.
\ov{9} It still remains unknown if Double Q-learning is provably faster than Q-learning, and this paper serves as an important step towards understanding Double Q-learning.
\ov{8} Given the current theoretical bounds, there is no evidence that DQL can surpass vanilla Q-learning in performance.
\ov{6} With the prevalence of Double Q-learning methods in various applications, it is of great importance to understand its theoretical guarantee.
\ov{7} Overall this paper is solid and very well written.
\\[2mm]

\end{longtable}

\end{document}